\documentclass{article}

\usepackage{arxiv}

\usepackage[utf8]{inputenc} 
\usepackage[T1]{fontenc}    
\usepackage{hyperref}       
\usepackage{url}            
\usepackage{booktabs}       
\usepackage{amsfonts}       
\usepackage{amsmath}
\usepackage{nicefrac}       
\usepackage{microtype}      
\usepackage{lipsum}		
\usepackage{graphicx}
\usepackage{natbib}
\usepackage{doi}
\usepackage{enumitem}
\usepackage{adjustbox}
\usepackage{tabularx}
\usepackage{booktabs}

\title{Bielik v3 Small: Technical Report}

\author{
 \textbf{Krzysztof Ociepa\textsuperscript{1,4}},
 \textbf{Łukasz Flis\textsuperscript{1,2}},
 \\
 \textbf{Remigiusz Kinas\textsuperscript{1}},
 \textbf{Krzysztof Wróbel\textsuperscript{1,3,5}},
 \textbf{Adrian Gwoździej\textsuperscript{1, 2}}
\\
\\
 \textsuperscript{1}SpeakLeash,
 \textsuperscript{2}ACK Cyfronet AGH,
 \textsuperscript{3}Jagiellonian University,
 \textsuperscript{4}Azurro,
 \textsuperscript{5}Enelpol
}

\date{}

\renewcommand{\undertitle}{The Bielik LLM Team}
\renewcommand{\shorttitle}{Bielik v3}

\hypersetup{
pdftitle={Bielik v3},
pdfsubject={q-bio.NC, q-bio.QM},
pdfauthor={Krzysztof Ociepa, Łukasz Flis, Remigiusz Kinas, Krzysztof Wróbel, Adrian Gwoździej},
}

\begin{document}
\maketitle

\begin{abstract}
We introduce Bielik v3, a series of parameter-efficient generative text models (1.5B and 4.5B) optimized for Polish language processing. These models demonstrate that smaller, well-optimized architectures can achieve performance comparable to much larger counterparts while requiring substantially fewer computational resources. Our approach incorporates key innovations, including a custom Polish tokenizer (APT4) that significantly improves token efficiency and Adaptive Learning Rate that dynamically adjusts based on training progress. Trained on a meticulously curated corpus of 292 billion tokens spanning 303 million documents, these models excel across multiple benchmarks, including the Open PL LLM Leaderboard, Complex Polish Text Understanding Benchmark, Polish EQ-Bench, and Polish Medical Leaderboard. The 4.5B parameter model achieves results competitive with models 2-3 times its size, while the 1.5B model delivers strong performance despite its extremely compact profile. These advances establish new benchmarks for parameter-efficient language modeling in less-represented languages, making high-quality Polish language AI more accessible for resource-constrained applications.
\end{abstract}


\section{Introduction}
The rapid advancement in natural language processing (NLP) has led to the development of increasingly sophisticated language models that can understand and generate human-like text. These models have shown remarkable success in various linguistic tasks across multiple languages. However, the development of high-performing models for less-resourced languages remains a significant challenge due to the scarcity of large and diverse datasets and computational resources.

Several notable efforts have advanced Polish language modeling in recent years. TRURL 2 \cite{trurl}, a collection of fine-tuned Llama 2 models with 7 billion and 13 billion parameters, was trained on approximately 1 million conversational samples. Qra \cite{qra} models, comprising continuously pretrained architectures with 1, 7, and 13 billion parameters, leveraged 90 billion tokens of Polish data. More recently, PLLuM, developed by a consortium of Polish academic institutions, introduced models ranging from 8 billion to 70 billion parameters, created through continued pretraining of Llama and Mistral models on Polish corpora. While these initiatives have made important strides, they often face limitations in performance, versatility, or accessibility, frequently requiring significantly larger computational resources for comparable performance.

Building on our previous work with Bielik 7B v0.1 \cite{ociepa2024bielik7bv01polish} and Bielik 11B v2 \cite{ociepa2025bielik11bv2technical}, we introduce the Bielik v3 series of generative text models optimized specifically for Polish language processing. These models, with sizes of 1.5B and 4.5B parameters, represent a significant advancement in parameter-efficient language modeling. By adopting an innovative approach to model scaling and training, we demonstrate that smaller, well-optimized models can achieve performance comparable to much larger counterparts while requiring substantially fewer computational resources.

Our approach introduces several key technical innovations. First, we employ depth up-scaling to adapt Qwen2.5 models \cite{qwen2025qwen25technicalreport}, replacing the original tokenizer with a custom-developed Polish tokenizer (APT4) that significantly improves token efficiency for Polish texts. Second, we utilize Adaptive Learning Rate, which dynamically adjusts the learning rate based on training progress and context length. These techniques, combined with comprehensive training on a diverse corpus of 292 billion tokens across 303 million documents, enable the Bielik v3 models to achieve remarkable performance despite their compact size.

Our evaluation demonstrates that Bielik v3 models outperform many larger models across various benchmarks, including the Open PL LLM Leaderboard, Complex Polish Text Understanding Benchmark (CPTUB), Polish EQ-Bench, and Polish Medical Leaderboard. Notably, the 4.5B parameter model achieves results competitive with models 2-3 times its size, while the 1.5B model delivers strong performance despite its extremely compact profile. This efficiency makes Bielik v3 particularly valuable for deployment in resource-constrained environments while maintaining high-quality Polish language capabilities.

In the following sections, we detail the model architecture and tokenizer design of Bielik v3, describe our comprehensive data preparation methodology, discuss the pre-training and post-training processes, and evaluate the models' performance across multiple benchmarks. We also analyze the models' limitations and potential biases. Our results demonstrate that Bielik v3 not only advances the state of Polish language understanding but also establishes new benchmarks for parameter-efficient language modeling in less-represented languages.

\section{Model and Tokenizer}
In this section, we introduce the model design and tokenizer, presenting architectural decisions and configurations.
resources
\subsection{Model Architecture}

\begin{table}[h]
\centering
\begin{tabular}{lll}
\toprule
Parameter & 1.5B & 4.5B \\
\midrule
Layers & 32 & 60 \\
Model Dimension & 1536 & 2048 \\
Attention Heads & 12 & 16 \\
Key/Value Heads & 2 & 2 \\
Head Size & 128 & 128 \\
Intermediate Size & 8960 & 11008 \\
Activation Function & SwiGLU & SwiGLU \\
Attention Bias & True & True \\
MLP Bias & True & True \\
Vocabulary Size & 32000 & 32000 \\
Positional Embeddings & RoPE ($\theta = 1000000$) & RoPE ($\theta = 1000000$) \\
Context Length & 8192 & 32768 \\
\bottomrule
\end{tabular}
\caption{Architecture details of the 1.5B and 4.5B parameter models.}
\label{tab:model-architecture}
\end{table}

\noindent Bielik v3 models are based on the Transformer architecture \cite{Vaswani2017AttentionIA}, with key parameters listed in Table \ref{tab:model-architecture}. The design integrates several advanced techniques to enhance performance and efficiency.

\noindent \textbf{Self-attention with causal masks} \cite{Vaswani2017AttentionIA} enables the model to assign varying importance to different parts of the input sequence. The causal mask ensures that the model only attends to preceding tokens, preserving the autoregressive property essential for language modeling.

\noindent \textbf{Grouped-query attention (GQA)} \cite{ainslie-etal-2023-gqa} reduces both computational complexity and memory usage while maintaining model quality. It achieves this by using fewer key-value heads than query heads, enabling more efficient handling of long sequences.

\noindent \textbf{SwiGLU activation function} \cite{Dauphin2016LanguageMW,shazeer2020gluvariantsimprovetransformer} combines the Swish activation function with Gated Linear Units (GLU), providing better performance and trainability than traditional activation functions such as ReLU.

\noindent \textbf{Rotary Positional Embeddings (RoPE)} \cite{SU2024127063} enhance the model's ability to capture relative token positions. Compared to absolute positional embeddings, RoPE supports better generalization to longer sequences and improves performance in tasks requiring positional sensitivity.

\noindent \textbf{Root Mean Square Layer Normalization (RMSNorm)} \cite{10.5555/3666122.3668105} normalizes activations within the network, offering greater training stability and slightly faster computation compared to standard Layer Normalization.

\noindent \textbf{Pre-normalization} involves applying layer normalization before the self-attention and feed-forward layers. This improves model convergence and overall performance.

\begin{figure}[ht]
\centering
\includegraphics[width=\columnwidth]{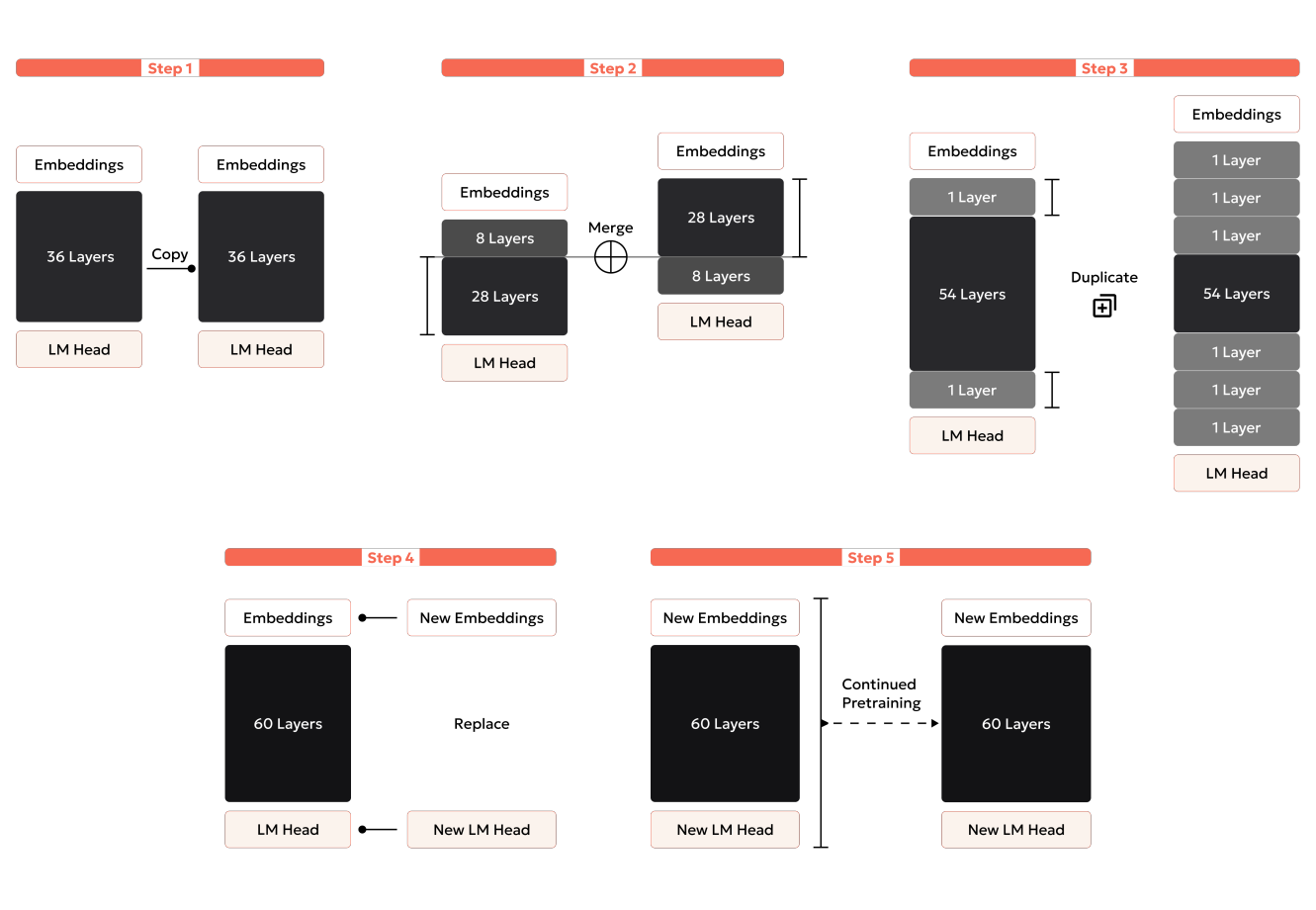}
\caption{Bielik 4.5B v3 model upscaling via Depth Up-Scaling ($n = 36$, $m = 8$, $s = 56$) with tokenizer replacement and outermost layer duplication.}
\label{fig:model-upscaling}
\end{figure}

The Bielik v3 models, with sizes 1.5B and 4.5B, are adapted from the Qwen2.5 1.5B and 3B models \cite{qwen2025qwen25technicalreport}. The models were scaled using the Depth Up-Scaling method \cite{kim2024solar107bscalinglarge}, the tokenizer was replaced, and further pretraining was conducted, as presented in Figure \ref{fig:model-upscaling}. The decision to build on an existing model rather than developing one from scratch was motivated by the desire to allocate resources efficiently. By focusing on the linguistic adaptation of an already high-performing model, we were able to optimize both time and computational resources. The Qwen2.5 models were selected due to their strong benchmark performance and permissive Apache 2.0 license.

\subsection{Tokenizer}

\begin{table*}[h]
\centering
\begin{tabular}{lll|lll|lll}
\toprule
                    &                 &                & \multicolumn{3}{c|}{Polish} & \multicolumn{3}{c}{English} \\
Tokenizer           & Vocab Size & Avg tokens & Tokens   & CpT    & TpW    & Tokens    & CpT    & TpW    \\
\midrule
APT3                & 31980           & 480            & 344      & 5.22   & 1.48   & 615       & 3.15   & 1.93   \\
APT4                & 32000           & 503            & 375      & 4.78   & 1.62   & 631       & 3.07   & 1.98   \\
Mistral v0.1        & 32000           & 578            & 747      & 2.40   & 3.22   & 408       & 4.75   & 1.28   \\
Qwen2.5             & 151665          & 499            & 625      & 2.87   & 2.69   & 373       & 5.19   & 1.17   \\
\bottomrule
\end{tabular}
\caption{Comparison of token count, characters per token (CpT), and tokens per word (TpW) for the preamble of the Constitution of the Republic of Poland in Polish and English, processed by various tokenizers: APT3 and APT4 (Polish-specific tokenizers), Mistral v0.1 and Qwen2.5 (multilingual tokenizers with limited Polish support).}
\label{tab:tokenizers-comparison}
\end{table*}

To enhance tokenization efficiency for the Polish language, we replaced the original tokenizer of the Qwen language model with our custom-developed Polish tokenizer, APT. This modification aimed to reduce the number of tokens required to represent input and output sequences, thereby enabling the model to handle longer contexts within its attention window and generate outputs more efficiently. Such improvements are particularly beneficial for smaller models, where token budget constraints are more pronounced.

One way to assess the effectiveness of the tokenization process is by analyzing the number of tokens generated for a given input. A lower token count generally indicates more efficient and faster text generation by the language model. The tokenizers used in the Mistral 7B and Qwen2.5 models were not specifically trained for the Polish language. Therefore, we decided to switch to a tokenizer trained primarily for Polish, with some support for English.

In addition to token count, we also considered how the tokenizer segments text—particularly whether it separates digits, punctuation, and special characters into distinct tokens, which can significantly impact the quality of generated responses. As a result, we ultimately chose to use our own APT4 tokenizer, which is a successor to the APT3 tokenizer from the Polish APT3 model \cite{AzurroAPT3Base1B}.

Adapting the model to the new tokenizer necessitated reinitializing the token embedding matrix to accommodate the altered vocabulary. We evaluated several embedding initialization methods:
\begin{itemize}
    \item Random Initialization: Assigns random vectors to new tokens, effectively requiring the model to learn embeddings from scratch, which can be inefficient and slow to converge.

    \item Frequency-based Vocabulary Transfer (FVT) \cite{yuan2022frequency}: Initializes embeddings for new tokens by averaging the embeddings of their constituent subword units from the original tokenizer, leveraging frequency information to inform the transfer.

    \item Linear Interpolation (aX + b): Applies a linear transformation to map embeddings from the source tokenizer's space to the target tokenizer's space, aiming to preserve relational structures between tokens.

    \item WECHSEL \cite{minixhofer2022wechsel}: Utilizes multilingual static word embeddings to identify semantically similar tokens between source and target vocabularies, initializing new token embeddings based on these similarities .

    \item FOCUS (Fast Overlapping Token Combinations Using Sparsemax) \cite{dobler2023focus}: Represents new tokens as sparse combinations of overlapping tokens from the source and target vocabularies, selected based on semantic similarity in an auxiliary embedding space .

    \item OFA (One For All) \cite{liu2023ofa}: Leverages external multilingual word embeddings to initialize unseen subword embeddings, facilitating efficient adaptation of pretrained models to new languages .

    \item RAMEN \cite{tran2020from}: Employs alignment techniques, such as bilingual dictionaries or cross-lingual embeddings, to map source language embeddings to target language tokens, aiding in the transfer of pretrained models to new languages.
\end{itemize}

After comparative analysis, we selected the FOCUS method for initializing embeddings corresponding to the APT tokenizer. FOCUS's approach of constructing new token embeddings as sparse combinations of semantically similar overlapping tokens proved effective in preserving the model's performance while accommodating the new tokenizer. To assess the efficacy of this tokenizer replacement and embedding initialization, we monitored the initial training loss of the model, providing insights into the immediate impact of these modifications on the model's learning dynamics.

After tokenizer change we duplicated the outermost layers of the model twice, allowing room for adaptation to the new embeddings. Next, we froze the entire model except for the duplicated layers and the embeddings, and trained it on a dataset containing 56 billion tokens. After this initial adaptation, we unfroze the entire model and continued training.

We selected the preamble of the Constitution of the Republic of Poland as the benchmark text because it effectively represents the style of Polish formal writing and is also available in an official English translation, enabling meaningful comparison. Table~\ref{tab:tokenizers-comparison} provides a detailed comparison of key metrics such as token count, characters per token (CpT), and tokens per word (TpW), demonstrating the relative performance of different tokenizers on both language versions of the preamble.

\section{Pre-training}

The primary objective of the pre-training phase was to enhance the model's proficiency in the Polish language, with an emphasis on both accuracy and fluency. To achieve this, we utilized a diverse collection of high-quality Polish texts. These materials underwent rigorous preprocessing and thorough quality evaluation to ensure the highest standards of training data, as shown in Tables~\ref{tab:classification-validation}, \ref{tab:classification-test}, and \ref{tab:shap}.

\subsection{Pre-training Data} \label{Pre-training-Data}
The pre-training of the Bielik v3 models involved constructing a novel, diverse, and high-quality dataset composed primarily of Polish-language texts. We leveraged resources from the SpeakLeash project \cite{speakleashorg}. Using metadata associated with each document—including topical information and various stylometric features—we selected 294 million documents from different datasets, ensuring both high quality and thematic diversity. These selected texts underwent comprehensive cleaning and quality evaluation, as described in Sections \ref{Data-Cleanup} and \ref{Quality-Evaluation}. 

Additionally, we excluded documents where scraping was technically permitted (i.e., not blocked by robots.txt) but where the terms and conditions explicitly prohibited use for training language models. Only documents meeting our stringent quality standards were retained and subsequently tokenized. This meticulous curation resulted in a Polish training corpus of 237 billion tokens.

To improve the model's adaptation to Polish while mitigating catastrophic forgetting \cite{Li2022OvercomingCF,pmlr-v199-ostapenko22a,ibrahim2024simplescalablestrategiescontinually}, we supplemented the dataset with English texts from the SlimPajama dataset \cite{cerebras2023slimpajama}, known for its diversity and quality.

To support the model's readiness for later training stages, we included the instruction dataset from Section~\ref{SFT-Data} as part of the pre-training corpus. Originally intended for supervised fine-tuning (SFT), this data contributed to a more seamless and efficient progression into the subsequent phases of training.

In total, the final training dataset comprised 292 billion tokens (303 million documents).

\subsubsection{Data Cleanup} \label{Data-Cleanup}
To enhance the quality of the documents, we applied a series of heuristics designed to remove corrupted or irrelevant content, anonymize personal data (including physical addresses, email addresses, phone numbers, and URLs), and resolve encoding or formatting issues. These steps produced cleaner, higher-quality texts that were subsequently subjected to further evaluation.

\subsubsection{Quality Evaluation} \label{Quality-Evaluation}

To develop the training dataset for text quality evaluation, we manually selected and annotated documents, categorizing them into three quality classes: \textbf{HIGH}, \textbf{MEDIUM}, and \textbf{LOW}. The \textbf{HIGH} class represents superior-quality documents, \textbf{LOW} denotes poor-quality texts, and \textbf{MEDIUM} encompasses documents whose quality is ambiguous, falling between high and low standards. This nuanced classification approach addresses the inherent complexities in assessing textual quality.

The Bielik v3 dataset comprises \textbf{44\,344} training documents, \textbf{3\,000} test documents, and \textbf{1\,000} validation documents.
In addition to real‑world texts, we introduced synthetic samples designed to probe the classifier's sensitivity to a wide range of quality degradations.

\subsection*{Synthetic Data Categories}

\begin{description}[leftmargin=0.55cm, style=nextline]
  \item[\textbf{HIGH‑quality synthetics}]%
        Well‑structured and linguistically fluent Markdown documents, demonstrating high factual coherence, clarity of expression, and grammatical correctness in Polish. These examples represent ideal outputs generated under constrained decoding settings.

  \item[\textbf{MEDIUM‑quality synthetics}]%
        Passages that are locally fluent and structurally plausible but exhibit moderate issues such as inconsistent tense, mild repetition, or topic drift. They reflect borderline cases between usable and discardable outputs.

  \item[\textbf{LOW‑quality synthetics}]%
        Outputs generated under high‑temperature sampling (e.g., temperature > 1.1), often formatted correctly in Markdown, but semantically incoherent, factually incorrect, or disorganized. This category also includes examples showing characteristic LLM failures such as looping or hallucinated content despite visually clean structure.  
        Additionally, it covers cases of extremely poor machine translation—especially from languages structurally distant from Polish, such as Chinese—where the resulting text, while superficially grammatical, is conceptually broken, mistranslated, or entirely nonsensical. These are now reliably detected and classified as low‑quality.
        Furthermore, we have introduced metrics aimed at detecting potentially looped or repetitive content, which sometimes occurred during text generation or editing using the Bielik v2 language model. These metrics effectively identify texts that may appear well-formatted and conceptually strong, but contain signs of excessive repetition or looping, raising concerns about content quality despite their otherwise high surface coherence.
\end{description}

These synthetic examples serve as both strong positives and strong negatives, enabling the model to move beyond surface‑level features and develop greater sensitivity to subtle semantic and structural failures.

\subsection*{Stylometric Feature Set}

To support classification, we designed an expanded set of stylometric features, aimed at capturing surface and deep characteristics of text typical for LLM-generated output. While the Bielik v2 classifier relied on 150 stylometric and Markdown-aware descriptors, Bielik v3 increases this to 200, introducing novel features tailored to detect generation artefacts, degraded machine translation, and formatting inconsistencies.

\begin{description}[leftmargin=0.55cm, style=nextline]
  \item[\textbf{Lexical richness and repetition}]%
        Features such as \textit{unique\_word\_count}, \textit{hapax\_legomena\_ratio}, and \textit{looping\_suspicion} help separate repetitive from genuinely varied prose.

  \item[\textbf{Diacritic and encoding hygiene}]%
        Counters of Polish diacritics, the \textit{replacement\_char\_ratio}, and related metrics expose corrupted character conversions.

  \item[\textbf{Sentence‑ and line‑level structure}]%
        Ratios of interrogative sentences or single‑word lines flag unusual formatting that often correlates with low factual quality.

  \item[\textbf{Readability indices}]%
        The \textit{lix} and \textit{rix} scores provide continuous proxies for extremely verbose or overly terse fragments.

  \item[\textbf{NER‑based coherence}]%
        The distribution of entity types (\textit{person}, \textit{organisation}, \textit{location}, \textit{miscellaneous}) helps detect hallucinations or long stretches of proper nouns without context.

  \item[\textbf{Morphosyntactic diversity}]%
        Features measuring variation in case, tense, and mood penalise unnaturally uniform or erratically shifting narratives.

  \item[\textbf{Part‑of‑speech–weighted top‑word ratios}]%
        Balances of nouns, verbs, and adjectives among the most frequent tokens reveal content focus versus padding.
\end{description}

Feature extraction and modelling
We compute these features with an extended StyloMetrix‑inspired pipeline \cite{okulska2023stylometrixopensourcemultilingualtool}, augmented with Unicode handling and a dependency parser for the new descriptors. Among the classifiers tested, XGBoost again achieved the highest macro‑F1 on the validation split; detailed feature‑importance rankings and ablations are reported in Tables 3 and 4.

The performance of the Bielik v3 quality classifier was rigorously evaluated on a held-out validation set. The model achieved an overall \textbf{accuracy of 95\%} and a \textbf{macro-average F1-score of 0.85}, demonstrating strong and balanced performance across all three quality categories.

Notably, the classifier reached an F1-score of \textbf{0.97} for both the HIGH and LOW classes, indicating excellent precision and recall in distinguishing clearly defined outputs. Despite lower recall in the MEDIUM category (\textbf{0.51}), which reflects its inherently ambiguous nature, the model maintains reliable boundaries between high- and low-quality texts—an essential requirement for downstream data curation.

Quality threshold for corpus construction
A manual audit of 1000 validation documents confirmed that a predicted probability P(HIGH) > 0.50 and P(MEDIUM) > 80 remains an effective cutoff. Documents below this threshold are excluded from the instruction‑tuning corpus used in downstream Bielik v3 training.

With the larger dataset, purpose‑built synthetic adversaries, and a richer 200‑dimensional feature vector, the Bielik v3 quality classifier more precisely detects both overt and subtle degradations, providing a robust filter for large‑scale LLM data.

\paragraph{Best model configuration.}
The best-performing model, \texttt{XGB\_RegL1L2\_d8\_n250\_lr008\_a02\_l05}, is an \texttt{XGBoostClassifier} trained with the following hyperparameters: 
\texttt{n\_estimators = 250}, \texttt{max\_depth = 8}, \texttt{learning\_rate = 0.08}, \texttt{subsample = 0.75}, 
\texttt{colsample\_bytree = 0.8}, \texttt{reg\_alpha = 0.2}, \texttt{reg\_lambda = 0.5}. The model uses the \texttt{multi:softprob} objective, 
was trained with \texttt{eval\_metric = mlogloss}, and a fixed \texttt{random\_state = 42}. Label encoding was disabled via \texttt{use\_label\_encoder = False}.

\begin{table}[htbp]
\centering
\begin{adjustbox}{max width=\textwidth}
\begin{tabularx}{\textwidth}{lccc}
\toprule
\textbf{Model} & \textbf{Val F1 (macro)} & \textbf{Val F1 (weighted)} & \textbf{Val Accuracy} \\
\midrule
XGB\_RegL1L2\_d8\_n250\_lr008\_a02\_l05 & 0.852303 & 0.941526 & 0.946 \\
XGB\_AggressiveSubsample\_d8\_n300\_lr007 & 0.841951 & 0.938001 & 0.943 \\
XGB\_RegL1\_d7\_n200\_lr01\_a05 & 0.841951 & 0.938001 & 0.943 \\
CatBoost & 0.841948 & 0.937998 & 0.943 \\
XGB\_RegL2\_d6\_n300\_lr005\_l1 & 0.838424 & 0.937647 & 0.943 \\
XGBoost\_nEstimators500\_maxDepth6\_lr007\_gamma15 & 0.836765 & 0.936393 & 0.941 \\
XGB\_StrongRegCombo\_d7\_n400\_lr003\_g2\_mcw4\_a01\_l01 & 0.834315 & 0.933820 & 0.939 \\
XGBoost\_nEstimators400\_maxDepth3\_lr025\_gamma05 & 0.832164 & 0.934450 & 0.938 \\
XGB\_VeryDeep\_LowLR\_d15\_n200\_lr005\_reg & 0.830983 & 0.934794 & 0.941 \\
HistGradientBoosting & 0.828799 & 0.933364 & 0.938 \\
XGBoost\_nEstimators180\_maxDepth12\_lr004\_minChild5 & 0.828858 & 0.931842 & 0.938 \\
XGBoost\_nEstimators100\_maxDepth6\_lr01 & 0.823295 & 0.931375 & 0.937 \\
XGBoost\_nEstimators200\_maxDepth8\_lr005 & 0.825671 & 0.931104 & 0.937 \\
LightGBM & 0.825986 & 0.930186 & 0.934 \\
XGBoost & 0.817084 & 0.927873 & 0.933 \\
MLP\_hidden100\_relu\_adam & 0.806844 & 0.913972 & 0.915 \\
EBM & 0.799630 & 0.914568 & 0.921 \\
MLP\_hidden100\_50\_relu\_lbfgs & 0.794733 & 0.915277 & 0.919 \\
TabNet & 0.793772 & 0.917838 & 0.920 \\
MLP\_hidden50\_50\_tanh\_sgd & 0.796052 & 0.912717 & 0.917 \\
RandomForest\_nEstimators300\_maxDepth20\_minSamples5 & 0.777787 & 0.914178 & 0.926 \\
\bottomrule
\end{tabularx}
\end{adjustbox}
\vspace{6pt}
\caption{Comparison of model performance on the validation set}
\label{tab:model-comparison}
\end{table}

\begin{table}[htbp]
\centering
\begin{tabular}{lcccc}
\toprule
Class & Precision & Recall & F1-score & Support \\
\midrule
LOW & 0.95 & 0.98 & 0.97 & 461 \\
MEDIUM & 0.79 & 0.51 & 0.62 & 74 \\
HIGH & 0.95 & 0.98 & 0.97 & 465 \\
\midrule
\textbf{Accuracy} & \multicolumn{3}{c}{0.95} & 1000 \\
\textbf{Macro avg} & 0.90 & 0.82 & 0.85 & 1000 \\
\textbf{Weighted avg} & 0.94 & 0.95 & 0.94 & 1000 \\
\bottomrule
\end{tabular}
\vspace{6pt}
\caption{Classification Report (Validation) – Best Quality Classifier Model}
\label{tab:classification-validation}
\end{table}

\begin{table}[htbp]
\centering
\begin{tabular}{lcccc}
\toprule
Class & Precision & Recall & F1-score & Support \\
\midrule
LOW & 0.94 & 0.97 & 0.96 & 1384 \\
MEDIUM & 0.77 & 0.47 & 0.58 & 225 \\
HIGH & 0.95 & 0.98 & 0.97 & 1391 \\
\midrule
\textbf{Accuracy} & \multicolumn{3}{c}{0.94} & 3000 \\
\textbf{Macro avg} & 0.89 & 0.81 & 0.83 & 3000 \\
\textbf{Weighted avg} & 0.93 & 0.94 & 0.93 & 3000 \\
\bottomrule
\end{tabular}
\vspace{6pt}
\caption{Classification Report (Test) – Best Quality Classifier Model}
\label{tab:classification-test}
\end{table}

\begin{figure}[ht]
\centering
\includegraphics[width=\columnwidth]{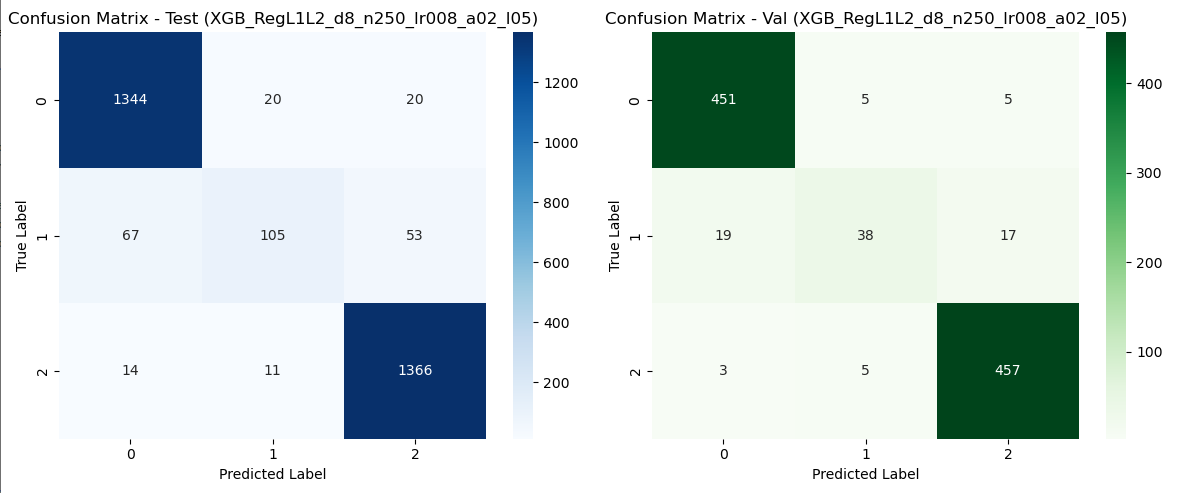}
\caption{Confusion matrix showing test and validation results for the XGBoost classifier.}
\label{fig:XGBoost-validation}
\end{figure}

\begin{table}[htbp]
\centering
\begin{tabular}{lc}
\toprule
\textbf{Feature} & \textbf{Mean Abs SHAP} \\
\midrule
oovs & 0.5726 \\
average\_lines & 0.2524 \\
stop\_word\_ratio & 0.2089 \\
polish\_diacritics\_per\_word & 0.1851 \\
non\_alpha\_word\_fractions & 0.1617 \\
rix & 0.1576 \\
short\_line\_ratio\_20 & 0.1524 \\
avg\_paragraph\_length & 0.1481 \\
lowercase\_ratio\_md & 0.1468 \\
single\_word\_line\_ratio & 0.1449 \\
colons\_per\_sentence & 0.1321 \\
not\_allowed\_chars\_ratio & 0.1122 \\
special\_chars\_ratio\_md & 0.1003 \\
symbol\_to\_word\_ratio & 0.0873 \\
duplicate\_line\_ratio & 0.0860 \\
commas\_per\_sentence & 0.0828 \\
blank\_lines\_ratio & 0.0801 \\
polish\_diacritics\_ratio & 0.0799 \\
char\_ratio\_\_ & 0.0776 \\
emoticons & 0.0759 \\
diacritics\_std\_dev & 0.0739 \\
single\_char\_ratio & 0.0704 \\
contextual\_word\_repetitions\_ratio & 0.0699 \\
overall\_uppercase\_ratio & 0.0679 \\
avg\_dependency\_tree\_depth & 0.0679 \\
short\_line\_ratio\_10 & 0.0662 \\
char\_ratio\_> & 0.0616 \\
\bottomrule
\end{tabular}
\vspace{6pt}
\caption{Top Features by Mean Absolute SHAP Value for quality classification}
\label{tab:shap}
\end{table}
\subsection{Category Classification: Results and Applications}

This section presents the performance evaluation of the text category classifier developed to automatically assign documents to one of 120 predefined categories. The model was trained and evaluated on a substantial dataset derived from Polish texts, with results summarized in Table~\ref{tab:category-classification-results}.

\subsubsection*{Dataset and Setup}
The dataset used for this task comprised a total of 58,294 documents. Following standard machine learning practice, the data was partitioned into a training set and a held-out test set. The training set consisted of 52,464 documents (approximately 90\% of the data), while the test set contained 5,830 documents (approximately 10\%).

A stratified splitting strategy was employed during partitioning to ensure that the proportional representation of each of the 120 categories was maintained in both subsets. This is crucial for reliable evaluation, especially given the large number of potentially imbalanced classes.

\subsubsection*{Modeling Approach}
The core classification model utilized a pipeline architecture. Textual data was first processed using a \texttt{CountVectorizer} to convert documents into numerical feature vectors based on word counts, limiting the vocabulary to the 15,000 most frequent terms. These counts were subsequently transformed into Term Frequency-Inverse Document Frequency (TF-IDF) representations using \texttt{TfidfTransformer}, capturing the relative importance of words across the corpus.

Classification was performed using a \texttt{Linear Support Vector Classifier} (LinearSVC), a robust and efficient algorithm for high-dimensional text data. To enable probability estimates and enhance performance, the LinearSVC was wrapped within a \texttt{CalibratedClassifierCV} using isotonic calibration (\texttt{method='isotonic'}) via 3-fold cross-validation on the training data.

\subsubsection*{Performance Evaluation}
The final trained pipeline was evaluated on the unseen test set (5,830 documents), demonstrating strong performance across the 120 categories.

The overall accuracy achieved was \textbf{94.63\%}, indicating that the vast majority of documents were correctly classified.

The high macro-average scores (all above 0.95) are particularly encouraging, suggesting that the classifier performs well across both common and rare categories.

Detailed per-class performance metrics, confusion matrix visualizations, and feature importance analyses were generated to examine specific strengths and weaknesses, identifying categories that may present greater classification challenges.


\begin{table}[htbp]
\centering
\begin{tabular}{lcccc}
\toprule
Metric & Precision & Recall & F1-score & Support \\
\midrule
\textbf{Macro avg} & 0.9533 & 0.9563 & 0.9543 & 5830 \\
\textbf{Weighted avg} & 0.9471 & 0.9463 & 0.9461 & 5830 \\
\bottomrule
\end{tabular}
\vspace{6pt}
\caption{Category Classification Performance on Test Set (N=5,830)}
\label{tab:category-classification-results}
\end{table}

\begin{figure}[!htbp]
\centering
\includegraphics[width=\columnwidth]{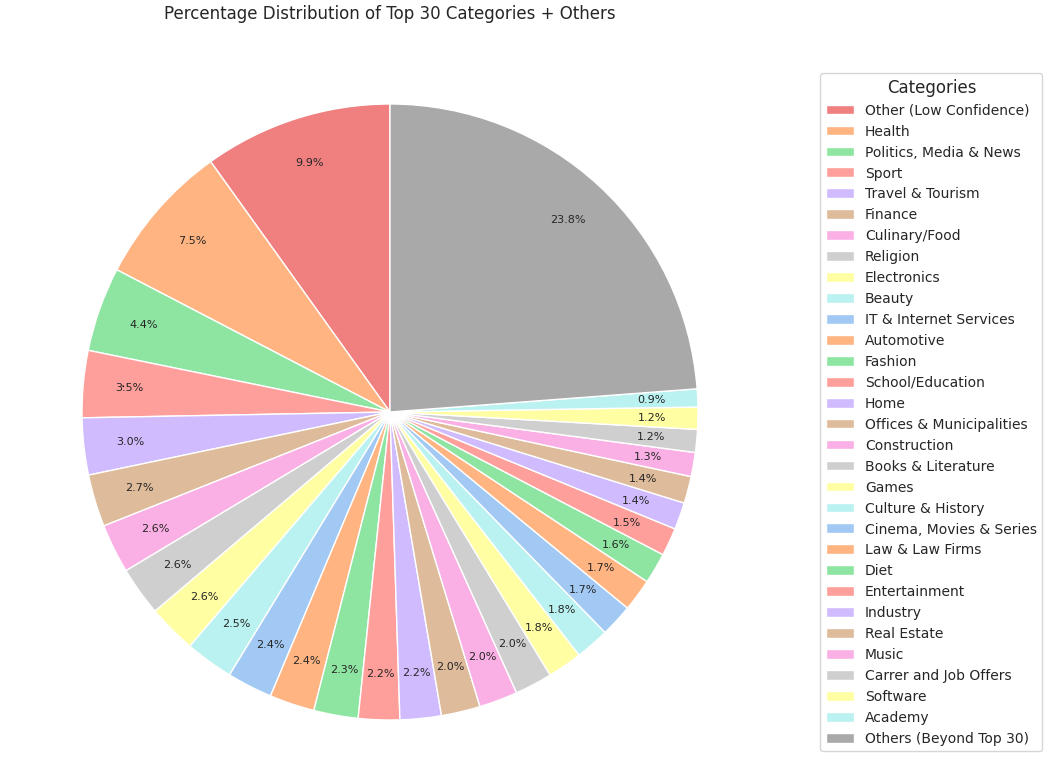}
\caption{Distribution of major thematic categories in the Polish text dataset ($\geq0.9\%$)}
\label{fig:distribution}
\end{figure}

\begin{table}[ht]
\centering
\begin{tabular}{|l|r|}
\hline
\textbf{Category} & \textbf{Count} \\
\hline
Other & 40,337,136 \\
Health & 30,432,887 \\
Politics, Media \& News & 18,066,594 \\
Sport & 14,308,164 \\
Travel \& Tourism & 12,168,387 \\
Finance & 11,182,892 \\
Culinary/Food & 10,537,704 \\
Religion & 10,499,588 \\
Electronics & 10,403,465 \\
Beauty & 10,337,631 \\
IT \& Internet Services & 9,707,525 \\
Automotive & 9,637,918 \\
Fashion & 9,481,582 \\
School/Education & 8,826,348 \\
Home & 8,810,490 \\
Offices \& Municipalities & 8,332,005 \\
Construction & 8,233,476 \\
Books \& Literature & 8,013,594 \\
Games & 7,418,487 \\
Culture \& History & 7,250,791 \\
Cinema, Movies \& Series & 6,984,202 \\
Law \& Law Firms & 6,945,003 \\
\hline
\multicolumn{2}{|c|}{--------} \\
\hline
Rescue Services & 132,048 \\
Diving & 130,770 \\
Lotteries & 129,202 \\
Bailiff Services & 128,669 \\
Currency Exchange & 117,801 \\
Plumbing & 107,582 \\
Taxi & 97,234 \\
Security Services & 59,082 \\
Postal Services & 52,035 \\
\hline
\end{tabular}
\vspace{6pt}
\caption{Distribution of text categories predicted by the classifier.}
\noindent
Note: The "Other" category includes texts where the classifier was uncertain and assigned a prediction with less than 
20\% confidence, as shown in Figure~\ref{fig:distribution}.
\label{tab:category_distribution}
\end{table}

\subsubsection*{Application and Future Directions}
The primary objective behind developing the category classifier, alongside the quality classifier, is to curate a dataset that is both thematically diverse and of the highest textual quality.

This dataset is intended for generating synthetic instruction data, leveraging varied and high-quality document collections. Ensuring both thematic breadth and content excellence will enable the creation of robust synthetic instructions, enhancing the effectiveness and generalization capabilities of downstream instruction-following models.

\section{Synthetic Data Generation for Instruction Tuning}

Building on the high-quality category and quality classification pipelines, we introduce a structured process for generating synthetic instruction datasets. This process ensures thematic diversity, high textual quality, and strategic reuse of imperfect data.

\subsection{Generation of High-Quality, Thematically Balanced QA Instructions}

The core of the synthetic data generation for QA task process involves creating instruction–response pairs from documents classified as \textbf{HIGH} quality and evenly distributed across thematic categories. 

Documents meeting the threshold of \textbf{P(HIGH) > 0.9} under the Bielik v3 quality classifier are selected. Category labels obtained from the thematic classifier ensure that sampling is performed in a balanced manner, avoiding overrepresentation of any single domain.

From these carefully curated inputs, task-oriented QA instructions are synthesized using controlled prompting techniques, ensuring coverage of a wide variety of instruction types, question styles, and domain-specific nuances. Emphasis is placed on factual coherence, clarity, and naturalness of the generated outputs, maintaining strict alignment with the standards of superior-quality human-authored instructions.

\subsection{Data Recycling: Improving Imperfect Texts for Inclusion}

Not all documents initially meet the strict inclusion criteria. Texts assessed as having \textbf{medium or borderline quality} undergo a dedicated recycling process before being incorporated into the base training corpus.

Specifically, documents categorized as:
\begin{itemize}
    \item \textbf{HIGH-quality texts} with a quality model confidence between 50\% and 70\%,
    \item \textbf{MEDIUM-quality texts} with high internal confidence scores,
\end{itemize}
are subjected to automated refinement using the Bielik v2.3 model.

The recycling stage primarily addresses superficial but systematic defects, including:
\begin{itemize}
    \item Spelling errors and typographical mistakes,
    \item Formatting irregularities and excessive or missing punctuation,
    \item Minor OCR artifacts or inconsistencies,
\end{itemize}
while preserving the underlying semantic content of the text.

Following refinement, these recycled documents are reassessed for quality. Only those meeting the revised thresholds are incorporated into the base model training datasets. This approach maximizes data utilization efficiency while maintaining high quality standards.

\subsection{Summary of the Synthetic Data Curation Strategy}

The synthetic data curation framework combines two key pillars:
\begin{itemize}
    \item \textbf{Selective instruction generation} from top-tier documents, ensuring thematic diversity and linguistic excellence.
    \item \textbf{Strategic data recycling}, repairing and upgrading moderately degraded documents to salvage valuable information without compromising quality.
\end{itemize}

By combining rigorous filtering and thematic balancing for instruction generation, and quality-preserving data recycling for the base training corpus, we construct datasets optimized respectively for instruction tuning and foundational model training, ensuring broad coverage and high factual reliability.

\section{Post-training}
Upon completing the pre-training phase, we transitioned to the post-training phase, which focused on further improving the model's performance across several domains, including coding, mathematics, logical reasoning, and instruction following.

\subsection{Supervised Fine-Tuning}
To better understand human behavior, the first step involves supervised fine-tuning (SFT), where a pretrained language model is adapted using dialogue-style datasets consisting of prompts and corresponding replies. The subsequent sections provide a detailed overview of the methods applied during this training process.

\subsubsection{Masked Tokens}
We introduced a masked token strategy where the loss function is applied selectively to certain parts of the output. Specifically, we masked the user instructions and control tokens from contributing to the loss \cite{shi2024instructiontuninglossinstructions}. This approach ensures that training focuses exclusively on the meaningful content tokens, avoiding unwanted optimization signals from non-content parts.

\subsubsection{Adaptive Learning Rate} \label{ALR}
Since instruction lengths vary widely, the number of tokens influencing the loss computation can fluctuate. To maintain consistent training dynamics, we implemented an adaptive learning rate (ALR) \cite{ociepa2024bielik7bv01polish}, scaling the base learning rate (LR) according to the square root of the ratio between the current batch's token count (T) and a reference batch size (BS):
\begin{equation}
\text{ALR} = \text{LR} \cdot \sqrt{\frac{\text{T}}{\text{BS}}}
\end{equation}

\subsection{Supervised Fine-Tuning Data}  \label{SFT-Data}

The dataset of instructions and dialogues in the Polish language, created for the Bielik 11B v2 release, was cleaned by removing low-quality samples and those containing various errors and defects. Some existing instructions were also regenerated, and completely new instructions were created using the Bielik 11B v2.3 model. The resulting dataset used for training included over 19 million instructions, totaling more than 13 billion tokens.

\subsection{Supervised Fine-Tuning Hyperparameters}

Training employed the AdamW optimizer with parameters $\beta_1 = 0.9$, $\beta_2 = 0.95$, and a weight decay of 0.05. The learning rate followed a cosine decay schedule, beginning at $7 \times 10^{-6}$ and gradually decreasing to $6 \times 10^{-7}$, with a warmup phase of 50 iterations.

We used a global batch size of 128, with each local batch consisting of a single sample. Gradient clipping was performed with a maximum norm of 1.0, and mixed-precision training was enabled using the \texttt{bfloat16} format.

To improve efficiency, we applied sample packing, combining multiple dataset samples into a single sequence until reaching the maximum allowed sequence length. The model was trained for 1.2 epochs with a maximum context length of 8,192 tokens.

\subsection{Preference Learning}

\subsubsection{Preference Training Methods}
In our exploration of post-training methodologies for aligning Bielik v3 model with human preferences, we conducted extensive evaluations of various preference optimization techniques. Building upon established methods such as Direct Preference Optimization (DPO), its penalized variant DPO-P, and Odds Ratio Preference Optimization (ORPO), we introduced and assessed a novel approach: Simple Preference Optimization (SimPO).

\textbf{Direct Preference Optimization (DPO)} simplifies the reinforcement learning from human feedback (RLHF) paradigm by eliminating the need for an explicit reward model. It directly optimizes the policy to prefer responses aligned with human preferences. 

\[
\mathcal{L}_{\text{DPO}}(\pi_\theta; \pi_{\text{ref}}) = - \mathbb{E}_{(x, y_{\text{w}}, y_{\text{l}}) \sim \mathcal{D}} \left[
\log \sigma \left(
\beta \left(
\log \frac{\pi_\theta(y_{\text{w}} \mid x)}{\pi_{\text{ref}}(y_{\text{w}} \mid x)} - \log \frac{\pi_\theta(y_{\text{l}} \mid x)}{\pi_{\text{ref}}(y_{\text{l}} \mid x)}
\right)
\right)
\right]
\]

Here, \( x \) represents the input prompt, \( y_{\text{w}} \) and \( y_{\text{l}} \) are the preferred and less preferred responses, respectively, \( \pi_\theta \) denotes the model's policy, \( \pi_{\text{ref}} \) is the reference policy, \( \beta \) is a scaling parameter, and \( \sigma \) is the sigmoid function. DPO addresses the complexity of RLHF by providing a stable and computationally efficient alternative that directly incorporates preference data into the training process.

\textbf{DPO with Penalty (DPO-P)} extends DPO by introducing a penalty term to account for uncertainty in preference data, mitigating overfitting to noisy or ambiguous annotations. This penalization adjusts the loss function to reduce the influence of uncertain samples, enhancing the robustness of the model to imperfect preference data. The DPO-P loss function is formulated as:

\[
\mathcal{L}_{\text{DPOP}}(\pi_\theta; \pi_{\text{ref}}) = - \mathbb{E}_{(x, y_{\text{w}}, y_{\text{l}}) \sim \mathcal{D}} \left[
\log \sigma \left(
\beta \left(
\log \frac{\pi_\theta(y_{\text{w}} \mid x)}{\pi_{\text{ref}}(y_{\text{w}} \mid x)} - \log \frac{\pi_\theta(y_{\text{l}} \mid x)}{\pi_{\text{ref}}(y_{\text{l}} \mid x)}
\right)
- \lambda \cdot \max\left(0, \log \frac{\pi_{\text{ref}}(y_{\text{w}} \mid x)}{\pi_\theta(y_{\text{w}} \mid x)} \right)
\right)
\right]
\]

Here, \( \lambda \) is a weighting factor for the penalty term. DPO-P addresses the challenge of aligning models without the overhead of additional reference models or complex training procedures.

\textbf{Odds Ratio Preference Optimization (ORPO)} ORPO integrates preference alignment into the supervised fine-tuning (SFT) phase by incorporating an odds ratio-based penalty. This approach eliminates the need for a separate reference model and simplifies the training pipeline. The ORPO loss function is formulated as:

\[
\mathcal{L}_{\text{ORPO}} = \mathcal{L}_{\text{NLL}} + \lambda \cdot \log \left( \frac{\pi_\theta(y_{\text{w}} \mid x)}{\pi_\theta(y_{\text{l}} \mid x)} \right)
\]

Where \( \mathcal{L}_{\text{NLL}} \) is the negative log-likelihood loss, and \( \lambda \) is a weighting factor for the odds ratio term. ORPO addresses the challenge of aligning models without the overhead of additional reference models or complex training procedures.

\textbf{Simple Preference Optimization (SimPO)} further streamlines preference optimization by utilizing the average log-probability of a sequence as an implicit reward, removing the necessity for a reference model. It introduces a target reward margin \( \gamma \) to enhance the separation between preferred and less preferred responses. The SimPO objective is defined as:

\[
\mathcal{L}_{\text{SimPO}} = -\mathbb{E}_{(x, y_{\text{w}}, y_{\text{l}}) \sim \mathcal{D}} \left[ \log \sigma\left( \beta \left( \frac{1}{|y_{\text{w}}|} \log \pi_\theta(y_{\text{w}} \mid x) - \frac{1}{|y_{\text{l}}|} \log \pi_\theta(y_{\text{l}} \mid x) - \gamma \right) \right) \right]
\]

Here, \( |y| \) denotes the length of the response, ensuring length normalization, and \( \gamma \) serves as the target reward margin. SimPO addresses the computational and memory inefficiencies associated with reference models, offering a more efficient and scalable solution for preference alignment. 

After conducting extensive evaluations across multiple Polish benchmarks—including we observed that the penalized variant of Direct Preference Optimization (DPO-P) consistently outperformed other methods such as DPO, ORPO, and SimPO. Despite the simplicity and computational efficiency offered by SimPO, DPO-P demonstrated superior alignment with human preferences, particularly in tasks requiring nuanced reasoning and factual accuracy.

\subsubsection{Preference Dataset}
Compared to the previous Bielik 11B v2 release, for the final phase of post-training with reinforcement learning, we introduced several key enhancements to our Polish preference instruction dataset and response generation pipeline. First, we significantly expanded the dataset to 126,000 instructions (only in Polish language), enriching its diversity and complexity. We also broadened the range of alignment categories to include function calling and tool calling tasks, while substantially increasing the volume of instructions focused on reasoning and mathematics. In addition, we incorporated a large number of conversational instructions to better reflect realistic interaction patterns. In the generation of preferred and rejected responses, we used a broader set of language models, notably including DeepSeek-V3-0324, alongside models used previously. Despite these improvements, we maintain a similar approach to instruction creation and curation, combining manual authoring, perturbation-based enhancement, rigorous deduplication, quality evaluation with reward metamodels, and manual inspection. These refinements ensure that the dataset not only grows in scale but also in quality, better supporting downstream alignment training.

\subsubsection{DPO-Positive Hyperparameters}

For DPO-Positive (DPO-P) training, we set the loss function parameters to $\beta = 0.1$ and $\lambda = 2.5$, in accordance with best practices aimed at stabilizing preference-based optimization and preserving the quality of preferred outputs.

The optimization was carried out using the AdamW algorithm, configured with $\beta_1 = 0.9$, $\beta_2 = 0.95$, and no weight decay. A fixed learning rate of $5 \times 10^{-7}$ was used, preceded by 50 warmup steps. The total number of training steps amounted to 1,800.

Training was performed with a global batch size of 64, where each local batch consisted of a single example. Gradient clipping was applied with a maximum norm of 1.0. Mixed-precision training was employed using the bfloat16 format.

\subsection{Reinforcement Learning}
For Reinforcement Learning (RL) training stage we employed Group Relative Policy Optimization (GRPO) \cite{shao2024deepseekmath} to fine-tune our Bielik-4.5B-v3 language model, utilizing a curated dataset of 12,000 Polish mathematical problems with verifiable solutions (RLVR). This subset was selected from a larger corpus of approximately 100,000 Polish-language math problems developed during the Bielik v3 project. The training process was conducted on an 8×GPU H100 cluster (Athena - Cyfronet) using the Volcano Engine Reinforcement Learning (VERL) framework \cite{sheng2024hybridflow} , which offers scalable and modular support for large language model (LLM) reinforcement learning workflows.

GRPO is a reinforcement learning algorithm designed to enhance reasoning capabilities in LLMs by addressing limitations found in traditional methods like Proximal Policy Optimization (PPO). Unlike PPO, which relies on a separate value function (critic) to estimate the expected return, GRPO eliminates the need for this component, thereby reducing memory consumption and computational complexity. Instead, GRPO evaluates multiple responses generated for the same prompt, computes their rewards using a reward model, and calculates the advantage of each response relative to the group's average reward. This group-based advantage estimation allows the model to update its policy by favoring responses that outperform the average, leading to more stable and efficient training.

In GRPO, for each input prompt $q$, the model generates a group of $G$ responses $\{a_1, a_2, \ldots, a_G\}$. Each response $a_i$ is evaluated using a reward function $R(q, a_i)$, which assesses the quality of the response. The mean $\mu$ and standard deviation $\sigma$ of the rewards within the group are computed as follows:

$$
\mu = \frac{1}{G} \sum_{i=1}^{G} R(q, a_i), \quad \sigma = \sqrt{\frac{1}{G} \sum_{i=1}^{G} (R(q, a_i) - \mu)^2}
$$

The advantage $A_i$ of each response $a_i$ is then calculated by normalizing its reward relative to the group's statistics:

$$
A_i = \frac{R(q, a_i) - \mu}{\sigma}
$$

This group-relative advantage estimation allows the model to identify which responses are better or worse compared to others in the same group, without requiring an explicit value function.

The policy is updated by maximizing the following objective function, which incorporates the advantage and a clipping mechanism to ensure stable updates:

$$
\mathcal{L}(\theta) = \frac{1}{G} \sum_{i=1}^{G} \min \left( r_i(\theta) A_i, \text{clip}(r_i(\theta), 1 - \epsilon, 1 + \epsilon) A_i \right)
$$

Here, $r_i(\theta) = \frac{\pi_\theta(a_i \mid q)}{\pi_{\theta_{\text{old}}}(a_i \mid q)}$ is the probability ratio between the new and old policies, and $\epsilon$ is a hyperparameter that controls the extent of clipping.

Additionally, GRPO incorporates a Kullback-Leibler (KL) divergence penalty to prevent the updated policy from deviating too much from a reference policy $\pi_{\text{ref}}$, typically the pre-trained model before fine-tuning. The KL penalty is added to the loss function as follows:

$$
\mathcal{L}_{\text{total}}(\theta) = \mathcal{L}(\theta) - \beta \cdot \text{KL}[\pi_\theta \parallel \pi_{\text{ref}}]
$$

Where $\beta$ is a coefficient that balances the importance of the KL penalty.

By leveraging group-relative advantages and eliminating the need for a value function, GRPO offers a more efficient and stable approach to fine-tuning LLMs, particularly in tasks that require complex reasoning, such as mathematical problem-solving.

The application of GRPO in our training regimen not only improved the model's performance on mathematical tasks but also enhanced its capabilities in other reasoning-intensive areas. This suggests that GRPO effectively strengthens the model's general reasoning abilities, making it a valuable approach for fine-tuning LLMs across diverse domains.

The optimization settings included a learning rate of $1 \times 10^{-6}$, a global batch size of 128 (with a local batch size of 16 per GPU). To ensure stable policy updates, KL divergence regularization was applied with a coefficient of 0.001, using the low-variance KL loss type.

\subsection{Model Merging}
Similar to the Bielik 11B v2 release, we applied a range of merging strategies to refine model quality; however, the leading method throughout the development of Bielik 11B v3 is the linear merging approach. This method served as the primary technique for improving model quality both after Supervised Fine-Tuning (SFT) and after Reinforcement Learning from Human Feedback (RLHF) stages. By consistently employing linear combinations of models at each phase, we ensured stable integration of improvements while maintaining the stylistic and functional consistency of the model outputs.

\section{Evaluation}

We evaluated the Bielik v3 models on several benchmarks to assess their performance across different language understanding and generation tasks, as detailed in the following subsections.

The models were evaluated on the following benchmarks:

\begin{itemize}
    \item \href{https://huggingface.co/spaces/speakleash/open_pl_llm_leaderboard}{Open PL LLM Leaderboard} \textbf{(Polish)}
    \item \href{https://huggingface.co/spaces/speakleash/polish_eq-bench}{Polish EQ-Bench} \textbf{(Polish)}
    \item \href{https://huggingface.co/spaces/speakleash/cptu_bench}{CPTUB Leaderboard} \textbf{(Polish)}
    \item \href{https://huggingface.co/spaces/speakleash/polish_medical_leaderboard}{Polish Medical Leaderboard} \textbf{(Polish)}
    \item \href{https://huggingface.co/spaces/sdadas/plcc}{Polish Linguistic and Cultural Competency Benchmark (PLCC)} \textbf{(Polish)}
    \item \href{https://huggingface.co/spaces/open-llm-leaderboard-old/open_llm_leaderboard}{Open LLM Leaderboard}
    \item \href{https://mixeval.github.io/}{MixEval}
    \item \href{https://gorilla.cs.berkeley.edu/leaderboard.html}{Berkeley Function-Calling Leaderboard}
\end{itemize}

\subsection{Open PL LLM Leaderboard} \label{Open-PL-LLM-Leaderboard}

The Open PL LLM Leaderboard, based on the Open LLM Leaderboard v1 \citep{open-llm-leaderboard-v1}, evaluates models on various NLP tasks, including: sentiment analysis, categorization, short answer question answering, and text classification, but does not test their conversational capabilities \citep{open-pl-llm-leaderboard,ociepa2024bielik7bv01polish}. The leaderboard utilizes the lm-evaluation-harness framework for model evaluation \citep{eval-harness}.

\paragraph{Tasks:}
\begin{itemize}
    \item \textbf{polemo2:} Sentiment analysis of online consumer reviews across four domains (medicine, hotels, products, university) with four-class labeling (positive, negative, neutral, ambiguous) \citep{kocon-etal-2019-multi}; metric: accuracy.
    \item \textbf{klej-ner:} Named entity recognition in sentences containing single-type entities, classifying into six categories (no entity, place, person, organization, time, geographical name) \citep{rybak-etal-2020-klej}; metric: accuracy.
    \item \textbf{8tags:} Topic classification of social media headlines into eight categories (film, history, food, medicine, motorization, work, sport, technology) \citep{dadas-etal-2020-evaluation}; metric: accuracy.
    \item \textbf{belebele:} Machine reading comprehension for question answering \citep{bandarkar-etal-2024-belebele}; metric: accuracy.
    \item \textbf{dyk:} Question answering based on human-annotated pairs from Wikipedia's "Did You Know" section \citep{marcinczuk2013open}; metric: binary F1.
    \item \textbf{ppc:} Text similarity assessment using manually labeled sentence pairs (exact paraphrases, close paraphrases, non-paraphrases) \citep{9945218}; metric: accuracy.
    \item \textbf{psc:} Summarization of news articles \citep{ogro:kop:14:lrec}; metric: binary F1.
    \item \textbf{cbd:} Text classification for cyberbullying and hate-speech detection \citep{ptaszynski2023expert}; metric: macro F1.
    \item \textbf{polqa:} Open-domain question answering from the "Jeden z dziesięciu" TV show, with and without context (abstractive QA/RAG) \citep{rybak-etal-2024-polqa-polish}; metric: accuracy, levenshtein.
    \item \textbf{poquad:} Context-based extractive question answering (QA/RAG) \citep{tuora2023poquad}; metric: levenshtein.
    \item \textbf{eqbench:} emotional intelligence benchmark \citep{paech2024eqbenchemotionalintelligencebenchmark}; metric: custom.
\end{itemize}

Most of the tasks are multiple-choice tests, which means that the model chooses the correct answer from a set of options.
They are implemented as two types of tests:
\begin{itemize}
    \item \textbf{Loglikelihood:} We choose the highest probability token from the given set, e.g., ABCD. These tests are suitable for base models.
    \item \textbf{Generate:} Model generates answer freely.
\end{itemize}

All tasks are evaluated in both 0-shot and 5-shot settings, with the average score across all tasks normalized by baseline scores.

It is important to note that PLLuM models are not included in this leaderboard, as they were trained on training portions of the tasks used in the benchmark (except for the Belebele and EQ-Bench tasks), unlike all other models present on the leaderboard, to the best of our knowledge. The authors of the datasets used in the benchmark are primarily PLLuM consortium members.

\begin{table*}[t]
\centering
\small
\begin{tabular}{lrr}
\toprule
\textbf{Model} & \textbf{Parameters (B)} & \textbf{Average} \\
\midrule
Qwen2.5-72B & 72.7 & 67.38 \\
Qwen2.5-32B & 32.8 & 66.73 \\
Qwen-72B & 72.7 & 66.02 \\
Qwen2.5-14B & 14.8 & 62.71 \\
Meta-Llama-3-70B & 70.6 & 62.07 \\
Qwen1.5-72B & 72.7 & 61.11 \\
Meta-Llama-3.1-70B & 70.6 & 60.87 \\
Mixtral-8x22B-v0.1 & 141.0 & 60.75 \\
Mistral-Small-24B-Base-2501 & 24.0 & 59.90 \\
Qwen1.5-32B & 32.8 & 58.71 \\
Bielik-11B-v2 & 11.2 & 58.14 \\
Qwen2.5-7B & 7.0 & 53.35 \\
EuroLLM-9B & 9.2 & 50.03 \\
Qwen-7B & 7.0 & 49.39 \\
SOLAR-10.7B-v1.0 & 10.7 & 47.54 \\
Mistral-Nemo-Base-2407 & 12.2 & 47.28 \\
internlm2-20b & 20.0 & 47.15 \\
\textbf{Bielik-4.5B-v3} & \textbf{4.8} & \textbf{45.47} \\
Qwen2.5-3B & 3.0 & 44.59 \\
Meta-Llama-3.1-8B & 8.0 & 43.77 \\
Meta-Llama-3-8B & 8.0 & 43.30 \\
Qwen1.5-72B & 72.3 & 39.51 \\
Mistral-7B-v0.3 & 7.0 & 38.88 \\
Mistral-7B-v0.2 & 7.0 & 38.81 \\
Qwen1.5-7B & 7.0 & 37.92 \\
Bielik-7B-v0.1 & 7.2 & 34.34 \\
Qra-13b & 13.0 & 33.90 \\
Llama-3.2-3B & 3.0 & 31.89 \\
Qwen2.5-1.5B & 1.5 & 31.83 \\
\textbf{Bielik-1.5B-v3} & \textbf{1.6} & \textbf{31.48} \\
Qra-7b & 7.0 & 16.60 \\
\bottomrule
\end{tabular}
\caption{Open PL LLM Leaderboard results for base models (5-shot evaluation)}
\label{tab:open-pl-llm-base}
\end{table*}

As shown in Table \ref{tab:open-pl-llm-base}, the Bielik-4.5B-v3 model achieves an impressive average score of 54.94, making it competitive with much larger models. This result is particularly noteworthy considering it outperforms several models with significantly more parameters, such as Qwen2.5-7B (53.35), EuroLLM-9B (50.03), and SOLAR-10.7B-v1.0 (47.54). The smaller Bielik-1.5B-v3 model achieves a score of 31.48, which is comparable to Qwen2.5-1.5B (31.83) and Llama-3.2-3B (31.89), despite its compact size.

\begin{table*}[t]
\centering
\small
\begin{tabular}{lrr}
\toprule
\textbf{Model} & \textbf{Parameters (B)} & \textbf{Average} \\
\midrule
Mistral-Large-Instruct-2411 & 123.0 & 69.84 \\
Meta-Llama-3.1-405B-Instruct-FP8 & 405.0 & 69.44 \\
Mistral-Large-Instruct-2407 & 123.0 & 69.11 \\
Qwen2.5-72B-Instruct & 72.7 & 67.92 \\
QwQ-32B-Preview & 32.8 & 67.01 \\
Llama-3.3-70B-Instruct & 70.6 & 66.40 \\
Qwen2-72B-Instruct & 72.7 & 65.87 \\
Bielik-11B-v2.3-Instruct & 11.2 & 65.71 \\
Bielik-11B-v2.2-Instruct & 11.2 & 65.57 \\
Meta-Llama-3.1-70B-Instruct & 70.6 & 65.49 \\
Bielik-11B-v2.1-Instruct & 11.2 & 65.45 \\
Mixtral-8x22B-Instruct-v0.1 & 141.0 & 65.23 \\
Bielik-11B-v2.0-Instruct & 11.2 & 64.98 \\
Meta-Llama-3-70B-Instruct & 70.6 & 64.45 \\
Qwen3-32B & 32.8 & 64.24 \\
Llama-4-Scout-17B-16E-Instruct & 109.0 & 64.21 \\
Bielik-11B-v2.5-Instruct & 11.2 & 63.95 \\
Mistral-Small-24B-Instruct-2501 & 24.0 & 62.97 \\
phi-4 & 14.7 & 62.57 \\
Qwen3-14B & 14.8 & 62.24 \\
Mistral-Small-Instruct-2409 & 22.2 & 61.41 \\
Qwen2.5-32B-Instruct & 32.8 & 61.21 \\
Qwen2.5-14B-Instruct & 14.8 & 59.91 \\
aya-23-35B & 35.0 & 56.37 \\
\textbf{
Bielik-4.5B-v3.0-Instruct} & \textbf{4.8} & \textbf{56.13} \\
Qwen3-8B & 8.2 & 55.78 \\
Qwen3-4B & 4.0 & 55.49 \\
Mistral-Nemo-Instruct-2407 & 12.2 & 55.27 \\
Qwen2.5-7B-Instruct & 7.6 & 54.93 \\
Mistral-7B-Instruct-v0.3 & 7.2 & 47.74 \\
Mistral-7B-Instruct-v0.2 & 7.2 & 45.95 \\
Bielik-7B-Instruct-v0.1 & 7.2 & 44.70 \\
Phi-4-mini-instruct & 3.8 & 43.30 \\
\textbf{
Bielik-1.5B-v3.0-Instruct} & \textbf{1.6} & \textbf{41.36} \\
Qwen2.5-3B-Instruct &	3.1 &	41.23 \\
Qwen3-1.7B & 2.0 & 38.34 \\
Mistral-7B-Instruct-v0.1 &	7.0 &	33.11 \\
Qwen2.5-1.5B-Instruct & 1.5 & 31.89 \\
\bottomrule
\end{tabular}
\caption{Open PL LLM Leaderboard results for instruction-tuned models (5-shot evaluation)}
\label{tab:open-pl-llm-instruct}
\end{table*}

The instruction-tuned models demonstrate substantial improvements over their base counterparts. As shown in Table \ref{tab:open-pl-llm-instruct}, Bielik-4.5B-v3.0-Instruct achieves a score of 56.13, outperforming Qwen2.5-7B-Instruct (54.93) and Mistral-Nemo-Instruct-2407 (55.27) despite having fewer parameters. Most impressively, Bielik-1.5B-v3.0-Instruct scores 41.36, exceeding the performance of Qwen2.5-3B-Instruct (41.23) with approximately half the parameters, and coming close to Phi-4-mini-instruct (43.30) which has more than twice the parameter count.

These results demonstrate the effectiveness of our training approach for the Bielik v3 models, which achieve remarkable parameter efficiency and strong performance on Polish language tasks. When considering the performance-to-parameter ratio, both the 1.5B and 4.5B models represent significant advancements in resource-efficient language modeling for Polish.

\subsection{Polish EQ-Bench}

The Polish Emotional Intelligence Benchmark, a localized Polish version of the original EQ-Bench \cite{paech2024eqbenchemotionalintelligencebenchmark}, evaluates language models' emotional intelligence capabilities across various dimensions of emotional understanding and response. This benchmark assesses models' ability to comprehend, interpret, and appropriately respond to emotionally complex situations in Polish language contexts.

\begin{table*}[t]
\centering
\small
\begin{tabular}{lrc}
\toprule
\textbf{Model} & \textbf{Parameters (B)} & \textbf{Score} \\
\midrule
Mistral-Large-Instruct-2407$^{\dagger}$ & 123.0 & 78.07 \\
Mistral-Large-Instruct-2411$^{\dagger}$ & 123.0 & 77.29 \\
Meta-Llama-3.1-405B-Instruct-FP8 & 405.0 & 77.23 \\
gpt-4o-2024-08-06 & Unknown & 75.15 \\
gpt-4-turbo-2024-04-09 & Unknown & 74.59 \\
Mistral-Small-Instruct-2409 & 22.2 & 72.85 \\
Llama-PLLuM-70B-chat & 70.6 & 72.56 \\
Meta-Llama-3.1-70B-Instruct & 70.6 & 72.53 \\
Bielik-11B-v2.5-Instruct & 11.2 & 72.00 \\
Qwen2-72B-Instruct & 72.7 & 71.23 \\
Meta-Llama-3-70B-Instruct & 70.6 & 71.21 \\
gpt-4o-mini-2024-07-18 & Unknown & 71.15 \\
Qwen2.5-32B-Instruct & 32.8 & 71.15 \\
Bielik-11B-v2.3-Instruct & 11.2 & 70.86 \\
Llama-3.3-70B-Instruct & 70.6 & 70.73 \\
Llama-PLLuM-70B-instruct & 70.6 & 69.99 \\
WizardLM-2-8x22B & 141.0 & 69.56 \\
Qwen2.5-14B-Instruct & 14.8 & 69.17 \\
Bielik-11B-v2.2-Instruct & 11.2 & 69.05 \\
Bielik-11B-v2.0-Instruct & 11.2 & 68.24 \\
glm-4-9b-chat & 9.0 & 61.79 \\
Mistral-Nemo-Instruct-2407 & 12.2 & 61.76 \\
Bielik-11B-v2.1-Instruct & 11.2 & 60.07 \\
EuroLLM-9B-Instruct & 9.2 & 54.10 \\
\textbf{Bielik-4.5B-v3.0-Instruct} & \textbf{4.8} & \textbf{53.58} \\
PLLuM-12B-chat & 12.2 & 52.26 \\
PLLuM-8x7B-nc-chat$^{\dagger}$ & 46.7 & 47.29 \\
Llama-PLLuM-8B-chat & 8.0 & 46.20 \\
Llama-3.2-3B-Instruct & 3.2 & 46.19 \\
PLLuM-8x7B-chat & 46.7 & 45.22 \\
PLLuM-8x7B-nc-instruct$^{\dagger}$ & 46.7 & 41.75 \\
PLLuM-8x7B-instruct & 46.7 & 39.55 \\
PLLuM-12B-instruct & 12.2 & 36.21 \\
Qwen2.5-3B-Instruct & 3.1 & 35.87 \\
PLLuM-12B-nc-chat$^{\dagger}$ & 12.2 & 35.41 \\
Llama-PLLuM-8B-instruct & 8.0 & 31.59 \\
Qwen2.5-1.5B-Instruct & 1.5 & 27.63 \\
Llama-3.2-1B-Instruct & 1.2 & 17.82 \\
gemma-1.1-2b-it & 2.5 & 16.47 \\
\textbf{Bielik-1.5B-v3.0-Instruct} & \textbf{1.6} & \textbf{13.88} \\
PLLuM-12B-nc-instruct$^{\dagger}$ & 12.2 & 13.11 \\
\bottomrule
\multicolumn{3}{l}{$^{\dagger}$Models with a non-commercial license.} \\
\end{tabular}
\caption{Polish EQ-Bench results for various models.}
\label{tab:pl-eq-bench}
\end{table*}

The results in Table~\ref{tab:pl-eq-bench} highlight the performance of Bielik v3 models on the emotionally nuanced Polish EQ-Bench. The Bielik-4.5B-v3.0-Instruct achieves a score of 53.58, which is particularly impressive for its parameter count. Despite having only 4.8B parameters, it outperforms several much larger models including PLLuM-12B-chat (52.26) and multiple PLLuM models with significantly more parameters. It also performs comparably to EuroLLM-9B-Instruct (54.10) with nearly half the parameters.

This efficiency is remarkable when considering the emotional intelligence capabilities achieved with substantially fewer parameters than comparable models. The performance gap between Bielik-4.5B-v3.0-Instruct and larger models like Bielik-11B-v2 variants (approximately 15-18 points difference) reflects the trade-offs between model size and performance, while demonstrating that even compact models can exhibit meaningful emotional intelligence capabilities.

The Bielik-1.5B-v3.0-Instruct model, with just 1.6B parameters, achieves a more modest score of 13.88, comparable to PLLuM-12B-nc-instruct (13.11) despite having only about 13\% of the parameters.

When considering the performance gradient across model sizes, we observe that the Bielik-4.5B-v3.0-Instruct achieves 76\% of the performance of our best Bielik-11B-v2.5-Instruct model (72.00) with only 43\% of the parameters. This efficient scaling pattern demonstrates the effectiveness of our training approach in balancing performance with computational efficiency across the Bielik model family.

These results demonstrate that the specialized training methodologies employed for Bielik v3 models enable them to achieve competitive performance on emotionally nuanced tasks despite their compact size, highlighting the effectiveness of the model architecture and training approach used for Polish language emotional intelligence capabilities.

\subsection{Complex Polish Text Understanding Benchmark (CPTUB)}

The Complex Polish Text Understanding Benchmark (CPTUB) \cite{cptub-leaderboard} is specifically designed to evaluate language models' proficiency in interpreting complex Polish texts. Unlike traditional tasks that focus on explicit meaning, CPTUB assesses the models' capacity to understand implied meanings and handle cognitively challenging questions. The benchmark comprises two main components:

\begin{itemize}
    \item \textbf{Implicatures}: Evaluates a model's ability to interpret implied meanings, including sarcasm, idiomatic expressions, and phraseological compounds. This component tests sensitivity to nuanced, context-dependent inferences through three subtasks:
    \begin{itemize}
        \item \textbf{Sentiment}: Correctly identifying the emotional tone beyond literal expressions
        \item \textbf{Language understanding}: Interpreting the underlying intentions of text authors
        \item \textbf{Phraseology}: Recognizing and explaining fixed or semi-fixed expressions whose meanings cannot be inferred from their individual components
    \end{itemize}
    \item \textbf{Tricky Questions}: Assesses a model's capability to address challenging questions characterized by logical puzzles, semantic ambiguity, logical inconsistencies, absurdity, and humor. This component specifically targets the model's reasoning skills and ability to avoid hallucinations when faced with ambiguous or nonsensical queries.
\end{itemize}

\begin{table*}[t]
\centering
\small
\begin{tabular}{lrccccccc}
\toprule
\textbf{Model} & \textbf{Params (B)} & \textbf{Overall} & \textbf{Implicatures} & \textbf{Senti-} & \textbf{Language} & \textbf{Phrase-} & \textbf{Tricky} \\
& & \textbf{Average} & \textbf{Average} & \textbf{ment} & \textbf{Understanding} & \textbf{ology} & \textbf{Questions} \\
\midrule
DeepSeek-R1 & 685.0 & 4.14 & 4.14 & 4.49 & 4.35 & 3.60 & 4.12 \\
Mistral-Large-Instruct-2411$^{\dagger}$ & 123.0 & 4.00 & 4.10 & 4.33 & 3.98 & 3.99 & 3.72 \\
Qwen2.5-72B-Instruct & 72.7 & 3.95 & 3.99 & 4.08 & 3.97 & 3.93 & 3.81 \\
Mistral-Large-Instruct-2407$^{\dagger}$ & 123.0 & 3.93 & 4.03 & 4.23 & 4.00 & 3.86 & 3.65 \\
Llama-4-Maverick-17B-128E-Instruct-FP8 & 402.0 & 3.93 & 3.99 & 4.39 & 4.11 & 3.48 & 3.76 \\
gemma-3-27b-it & 27.4 & 3.81 & 3.90 & 3.88 & 3.79 & 4.03 & 3.53 \\
Meta-Llama-3-70B-Instruct & 70.6 & 3.78 & 3.81 & 4.13 & 3.82 & 3.47 & 3.71 \\
Qwen2.5-32B-Instruct & 32.8 & 3.75 & 3.80 & 3.81 & 3.57 & 4.04 & 3.59 \\
Llama-4-Scout-17B-16E-Instruct-FP8 & 402.0 & 3.75 & 3.94 & 4.10 & 3.81 & 3.90 & 3.19 \\
Bielik-11B-v2.3-Instruct & 11.2 & 3.63 & 3.77 & 3.97 & 3.79 & 3.55 & 3.22 \\
Bielik-11B-v2.1-Instruct & 11.2 & 3.61 & 3.66 & 3.96 & 3.92 & 3.47 & 3.47 \\
Mixtral-8x22B-Instruct-v0.1 & 141.0 & 3.56 & 3.67 & 3.78 & 3.68 & 3.55 & 3.24 \\
Qwen2.5-14B-Instruct & 14.8 & 3.55 & 3.62 & 3.91 & 3.57 & 3.37 & 3.34 \\
Llama-PLLuM-70B-chat & 70.6 & 3.53 & 3.63 & 3.94 & 3.61 & 3.35 & 3.21 \\
Bielik-11B-v2.5-Instruct & 11.2 & 3.48 & 3.67 & 4.01 & 3.86 & 3.13 & 2.91 \\
Bielik-11B-v2.2-Instruct & 11.2 & 3.46 & 3.57 & 3.72 & 3.73 & 3.25 & 3.12 \\
\textbf{Bielik-4.5B-v3.0-Instruct} & \textbf{4.8} & \textbf{3.38} & \textbf{3.68} & \textbf{3.76} & \textbf{3.61} & \textbf{3.67} & \textbf{2.46} \\
Llama-PLLuM-70B-instruct & 70.6 & 3.33 & 3.56 & 3.78 & 3.63 & 3.26 & 2.63 \\
phi-4 & 14.7 & 3.30 & 3.50 & 3.72 & 3.54 & 3.24 & 2.72 \\
Bielik-11B-v2.0-Instruct & 11.2 & 3.26 & 3.61 & 3.97 & 3.75 & 3.13 & 2.20 \\
PLLuM-12B-nc-chat$^{\dagger}$ & 12.2 & 3.15 & 3.33 & 3.22 & 3.23 & 3.54 & 2.62 \\
PLLuM-12B-chat & 12.2 & 3.14 & 3.32 & 3.32 & 3.21 & 3.43 & 2.59 \\
PLLuM-8x7B-nc-instruct$^{\dagger}$ & 46.7 & 3.11 & 3.56 & 3.88 & 3.59 & 3.22 & 1.76 \\
PLLuM-12B-instruct & 12.2 & 3.09 & 3.49 & 3.71 & 3.17 & 3.59 & 1.90 \\
Qwen2.5-7B-Instruct & 7.62 & 3.07 & 3.23 & 3.56 & 3.03 & 3.10 & 2.58 \\
PLLuM-8x7B-nc-chat$^{\dagger}$ & 46.7 & 3.03 & 3.44 & 3.76 & 3.48 & 3.08 & 1.80 \\
Meta-Llama-3.1-8B-Instruct & 8.0 & 3.01 & 3.31 & 3.97 & 3.38 & 2.58 & 2.11 \\
PLLuM-8x7B-instruct & 46.7 & 3.01 & 3.51 & 3.59 & 3.47 & 3.46 & 1.51 \\
PLLuM-8x7B-chat & 46.7 & 3.01 & 3.41 & 3.44 & 3.45 & 3.35 & 1.78 \\
Meta-Llama-3-8B-Instruct & 8.0 & 3.00 & 3.17 & 3.33 & 3.15 & 3.04 & 2.48 \\
Llama-PLLuM-8B-chat & 8.0 & 2.92 & 3.14 & 3.13 & 2.93 & 3.36 & 2.25 \\
Bielik-7B-Instruct-v0.1 & 7.2 & 2.88 & 3.13 & 3.59 & 3.48 & 2.32 & 2.16 \\
Llama-PLLuM-8B-instruct & 8.0 & 2.82 & 3.20 & 3.24 & 2.90 & 3.46 & 1.66 \\
gemma-2-2b-it & 2.6 & 2.65 & 2.80 & 3.40 & 2.90 & 2.10 & 2.21 \\
Qwen2.5-3B-Instruct & 3.1 & 2.50 & 2.73 & 2.95 & 2.46 & 2.80 & 1.81 \\
\textbf{Bielik-1.5B-v3.0-Instruct} & \textbf{1.6} & \textbf{2.36} & \textbf{2.75} & \textbf{3.53} & \textbf{2.33} & \textbf{2.38} & \textbf{1.22} \\
Phi-4-mini-instruct & 3.8 & 2.17 & 2.46 & 2.69 & 2.43 & 2.25 & 1.30 \\
Llama-3.2-3B-Instruct & 3.2 & 2.00 & 2.26 & 2.76 & 2.30 & 1.72 & 1.22 \\
EuroLLM-1.7B-Instruct & 1.7 & 1.76 & 2.10 & 2.24 & 1.79 & 2.26 & 0.76 \\
Qwen2.5-1.5B-Instruct & 1.5 & 1.76 & 2.12 & 2.79 & 1.35 & 2.23 & 0.66 \\

\bottomrule
\multicolumn{8}{l}{$^{\dagger}$Models with a non-commercial license.} \\
\end{tabular}
\caption{Complex Polish Text Understanding Benchmark (CPTUB) results across different evaluation categories}
\label{tab:cptub}
\end{table*}

As shown in Table \ref{tab:cptub}, the Bielik v3 models demonstrate impressive performance on this challenging benchmark, particularly in relation to their parameter counts. The Bielik-4.5B-v3.0-Instruct model achieves an overall score of 3.38, which is remarkable for a model with only 4.8B parameters. This positions it in the same performance tier as significantly larger models, including several with over 10x the parameter count.

Several key observations can be made about the Bielik v3 models' performance:

\begin{enumerate}
    \item \textbf{Exceptional parameter efficiency}: Bielik-4.5B-v3.0-Instruct (3.38) outperforms phi-4 (3.30) despite having only about a third of the parameters (4.8B vs. 14.7B). It also surpasses all PLLuM models regardless of size, including PLLuM-8x7B variants with nearly 10x more parameters.
    
    \item \textbf{Strong implicature handling}: Bielik-4.5B-v3.0-Instruct shows particularly strong performance in implicatures (3.68), exceeding even some Bielik-11B-v2 variants and models like Mixtral-8x22B-Instruct-v0.1 (3.67). This suggests superior understanding of nuanced Polish linguistic features like idioms and contextual meaning.
    
    \item \textbf{Phraseology strength}: The Bielik-4.5B-v3.0-Instruct model scores 3.67 in phraseology, notably higher than many larger models including all Bielik-11B variants. This indicates exceptional ability to understand Polish idiomatic expressions and fixed phrases.
    
    \item \textbf{Sentiment analysis competence}: Both Bielik v3 models perform well in sentiment analysis, with the 1.5B model scoring 3.53 - higher than many larger models including Qwen2.5-3B-Instruct (2.95) despite having half the parameters.
    
    \item \textbf{Tricky questions challenges}: The area with the most room for improvement is in handling tricky questions, where Bielik-4.5B-v3.0-Instruct scores 2.46. This is consistent with the pattern seen across most models, as this category tests complex reasoning abilities that typically benefit from larger model scales.
\end{enumerate}

The performance of Bielik-1.5B-v3.0-Instruct is similarly impressive within its parameter class. At just 1.6B parameters, it achieves an overall score of 2.36, outperforming models with substantially more parameters like Phi-4-mini-instruct (2.17 with 3.8B parameters) and performing comparably to Qwen2.5-3B-Instruct (2.50 with 3.1B parameters). Its strong sentiment analysis score (3.53) is particularly noteworthy, matching or exceeding many models with 3-7x more parameters.

These results highlight the effectiveness of the specialized training methodologies employed for the Bielik v3 models, particularly the focus on Polish-specific data curation and the innovative techniques described in previous sections, as shown in Tables~\ref{tab:model-architecture} and~\ref{tab:tokenizers-comparison}. The models demonstrate that through careful optimization, even relatively small models can achieve competitive performance on complex linguistic tasks that traditionally favor much larger architectures.

\subsection{Polish Medical Leaderboard}

The Polish Medical Leaderboard evaluates language models on Polish Board Certification Examinations (Państwowy Egzamin Specjalizacyjny, PES) from years 2018-2022. This benchmark assesses models' medical knowledge and reasoning capabilities in a Polish-language medical context, using datasets from speakleash/PES-2018-2022, which is based on amu-cai/PES-2018-2022 \cite{pokrywka2024gpt4passes}.

\begin{table*}[t]
\centering
\small
\begin{tabular}{lrc}
\toprule
\textbf{Model} & \textbf{Parameters (B)} & \textbf{Average (\%)} \\
\midrule
Meta-Llama-3.1-405B-Instruct-FP8 & 405.0 & 69.20 \\
Mistral-Large-Instruct-2407$^{\dagger}$ & 123.0 & 64.28 \\
Qwen2.5-72B-Instruct & 72.7 & 63.89 \\
Meta-Llama-3.1-70B-Instruct & 70.6 & 61.75 \\
Qwen2-72B-Instruct & 72.7 & 61.35 \\
Meta-Llama-3-70B-Instruct & 70.6 & 57.51 \\
Qwen2.5-32B & 32.8 & 55.69 \\
Qwen2.5-32B-Instruct & 32.8 & 54.52 \\
Qwen2.5-14B-Instruct & 14.8 & 49.60 \\
Bielik-11B-v2.5-Instruct & 11.2 & 44.85 \\
GLM-4-9b-chat & 9.0 & 44.54 \\
Mistral-Small-Instruct-2409 & 22.2 & 43.60 \\
\textbf{Bielik-4.5B-v3.0-Instruct} & \textbf{4.8} & \textbf{43.55} \\ 
Bielik-11B-v2.3-Instruct & 11.2 & 43.26 \\
Bielik-11B-v2.1-Instruct & 11.2 & 43.16 \\
Bielik-11B-v2.2-Instruct & 11.2 & 43.05 \\
Qwen2.5-7B-Instruct & 7.6 & 42.69 \\
Bielik-11B-v2.0-Instruct & 11.2 & 41.53 \\
Meta-Llama-3.1-8B-Instruct & 8.0 & 40.60 \\
Mistral-Nemo-Instruct-2407 & 12.2 & 40.36 \\
Bielik-11B-v2 & 11.2 & 39.98 \\
Qwen2.5-3B-Instruct & 3.0 & 37.72 \\
\textbf{Bielik-1.5B-v3.0-Instruct} & \textbf{1.6} & \textbf{34.63} \\
Qwen2.5-1.5B-Instruct & 1.5 & 32.64 \\
Mistral-7B-Instruct-v0.3 & 7.0 & 31.24 \\
\bottomrule
\multicolumn{3}{l}{$^{\dagger}$Models with a non-commercial license.} \\
\end{tabular}
\caption{Polish Medical Leaderboard results (5-shot setting) showing model performance on Polish Board Certification Examinations.}
\label{tab:medical-leaderboard}
\end{table*}

\paragraph{Bielik v3's performance:} In the Polish Medical Leaderboard (Table~\ref{tab:medical-leaderboard}), the Bielik v3 models demonstrate impressive medical reasoning capabilities relative to their model size:
\begin{itemize}
    \item Bielik-4.5B-v3.0-Instruct achieves a score of 43.55\%, which is remarkably close to Bielik-11B-v2.5-Instruct (44.85\%) despite having less than half the parameters
    \item The smaller Bielik-1.5B-v3.0-Instruct scores 34.63\%, outperforming Qwen2.5-1.5B-Instruct (32.64\%) and significantly outperforming larger models like Mistral-7B-Instruct-v0.3 (31.24\%)
    \item Notably, Bielik-4.5B-v3.0-Instruct matches or outperforms several much larger models, including Mistral-Small-Instruct-2409 (43.60\%) which has 22.2B parameters
\end{itemize}

\paragraph{Performance context:} The benchmark highlights several important insights about Bielik v3's medical capabilities:
\begin{itemize}
    \item Both Bielik v3 models achieve strong parameter efficiency, with the 4.5B model performing at nearly the same level as models 2-3 times its size
    \item The models show effective cross-domain generalization from general Polish language understanding to specialized medical knowledge, despite not having domain-specific medical training
    \item The gap between Bielik v3 models and top-performing models like Meta-Llama-3.1-405B-Instruct (69.20\%) reflects the expected scaling relationship between model size and specialized domain knowledge
\end{itemize}

These results emphasize the effectiveness of the training methodologies employed for the Bielik v3 models, enabling strong performance on specialized domain knowledge even at smaller parameter counts. This makes the Bielik v3 models particularly valuable for practical applications where computational efficiency must be balanced with domain-specific performance.

\subsection{Polish Linguistic and Cultural Competency Benchmark (PLCC)}

The Polish Linguistic and Cultural Competency Benchmark (PLCC) \citep{dadas2025evaluatingpolishlinguisticcultural} is a specialized evaluation framework that tests language models' grasp of Polish cultural knowledge and context. Unlike conventional NLP assessments, PLCC delves deeper into cultural understanding through 600 carefully designed questions spanning six key domains: history, geography, culture \& tradition, art \& entertainment, grammar, and vocabulary.

The benchmark's questions probe models' familiarity with various aspects of Polish heritage, including cultural references, historical milestones, traditional customs, folklore, literary works, and contemporary popular culture. These elements are crucial for achieving authentic language comprehension that goes beyond mere text processing. The questions vary in complexity from widely known information to specialized regional knowledge, presented in both multiple-choice and open-ended formats that demand precise responses with specific facts, dates, names, or concepts.

\begin{table*}[hbt!]
\centering
\small
\begin{tabular}{lrc}
\toprule
\textbf{Model} & \textbf{Parameters (B)} & \textbf{Average Score (\%)} \\
\midrule
Gemini-2.5-Pro-Exp-03-25	& Unknown &	89.50 \\
DeepSeek-R1 & 685.0 & 76.00 \\
DeepSeek-v3-0324	& 685.0 &	71.00 \\
DeepSeek-v3		& 685.0 &69.17 \\
PLLuM-8x7B-nc-chat$^{\dagger}$ & 46.7 & 68.17 \\
Llama-3.1-Tulu-3-405B	&  405.0 &	63.83 \\
Bielik-11B-v2.2-Instruct & 11.2 & 63.00 \\
Bielik-11B-v2.3-Instruct & 11.2 & 62.17 \\
GPT-4.1-mini-2025-04-14	& Unknown &	62.17 \\
Bielik-11B-v2.1-Instruct & 11.2 & 61.00 \\
Llama-3.1-405B & 405.0 & 60.00 \\
PLLuM-12B-nc-chat$^{\dagger}$ & 12.2 & 59.50 \\
Llama-PLLuM-70B-chat & 70.6 & 58.50 \\
Llama-4-Maverick & 402.0 & 58.17 \\
Command-A-03-2025$^{\dagger}$ & 111.0 & 56.17 \\
Mistral-Large-2407$^{\dagger}$ & 123.0 & 54.17 \\
PLLuM-8x7B-chat & 46.7 & 54.17 \\
Mistral-Large-2411$^{\dagger}$ & 123.0 & 52.00 \\
WizardLM-2-8x22B & 141.0 & 51.50 \\
Qwen-Max & Unknown & 50.83 \\
Command-R-Plus-08-2024$^{\dagger}$ & Unknown & 50.17 \\
Mixtral-8x22B & 141.0 & 49.83 \\
Command-R-Plus-04-2024$^{\dagger}$ & Unknown & 49.33 \\
Llama-3.3-70B & 70.6 & 48.83 \\
Llama-3.1-70B & 70.0 & 47.83 \\
Gemma-3-27B & 27.4 & 47.33 \\
PLLuM-12B-chat & 12.2 & 47.00 \\
Bielik-7B-Instruct-v0.1 & 7.0 & 46.67 \\
Mistral-Small-3.1-24B-2503 & 24.0 & 43.33 \\
Llama-3.0-70B & 70.0 & 43.00 \\
Gemma-2-27B & 27.0 & 42.67 \\
\textbf{Bielik-4.5B-v3.0-Instruct} & \textbf{4.8} & \textbf{42.33} \\
Llama-4-Scout & 109.0 & 41.50 \\
EuroLLM-9B & 9.0 & 41.00 \\
Qwen-2.5-72B & 72.7 & 39.17 \\
Mistral-Small-24B-2501 & 24.0 & 39.00 \\
Llama-PLLuM-8B-chat & 8.0 & 38.50 \\
Qwen3-32B & 32.8 & 37.67 \\
Mixtral-8x7B & 46.7 & 35.33 \\
Qwen-2.5-32B & 32.8 & 30.50 \\
Qwen3-14B & 14.8 & 30.33 \\
Gemma-2-9B & 9.0 & 29.17 \\
Phi-4 & 14.7 & 29.17 \\
\textbf{Bielik-1.5B-v3.0-Instruct} & \textbf{1.6} & \textbf{27.50} \\
Qwen-2.5-14B & 14.8 & 26.67 \\
Qwen3-8B & 8.2 & 26.00 \\
Mistral-Nemo & 12.2 & 23.00 \\
Command-R-7B$^{\dagger}$ & 7.0 & 22.83 \\
Llama-3.1-8B & 8.0 & 22.67 \\
Mistral-7B-v0.3 & 7.2 & 21.83 \\
Ministral-8B & 8.0 & 20.67 \\
Qwen-2.5-7B & 7.0 & 17.67 \\
\bottomrule
\multicolumn{3}{l}{$^{\dagger}$Models with a non-commercial license.} \\
\end{tabular}
\caption{Polish Linguistic and Cultural Competency Benchmark (PLCC) results for open-source models. Closed proprietary models have been excluded from this comparison.}
\label{tab:plcc-scores}
\end{table*}

The PLCC results (Table~\ref{tab:plcc-scores}) reveal several insights about Bielik v3 models' cultural competency:

\paragraph{Parameter efficiency:} Bielik-4.5B-v3.0-Instruct achieves 42.33\% on PLCC despite its small size, outperforming several larger models including Qwen-2.5-14B (26.67\%) and Phi-4 (29.17\%). Similarly, Bielik-1.5B-v3.0-Instruct reaches 27.50\%, demonstrating strong performance for its compact parameter count.

\paragraph{Size-performance tradeoff:} While Bielik v3 models score lower than their larger Bielik v2 counterparts (Bielik-11B-v2.2-Instruct at 63.00\%), they provide viable alternatives for resource-constrained deployments while maintaining reasonable cultural knowledge capabilities.

\paragraph{Competitive small-model performance:} Bielik-4.5B-v3.0-Instruct shows competitive results compared to models with similar parameter counts, highlighting the effectiveness of our training methodology in preserving cultural knowledge despite parameter reduction.

\paragraph{Progressive improvements:} The performance gap between Bielik-1.5B-v3.0-Instruct (27.50\%) and Bielik-4.5B-v3.0-Instruct (42.33\%) demonstrates the benefits of additional parameters for cultural knowledge retention, while both maintain efficient footprints.

\subsection{Open LLM Leaderboard}

The Open LLM Leaderboard \citep{open-llm-leaderboard} evaluates models on various English language tasks, providing insights into the model's performance across different linguistic challenges.

\begin{table*}[t]
\centering
\small
\begin{tabular}{lccccccc}
\toprule
\textbf{Model} & \textbf{AVG} & \textbf{arc\_challenge} & \textbf{hellaswag} & \textbf{truthfulqa\_mc2} & \textbf{mmlu} & \textbf{winogrande} & \textbf{gsm8k} \\
\midrule
Qwen1.5-14B & 66.70 & 56.57 & 81.08 & 52.06 & 69.36 & 73.48 & 67.63 \\
Bielik-11B-v2 & 65.87 & 60.58 & 79.84 & 46.13 & 63.06 & 77.82 & 67.78 \\
Qwen-14B & 65.86 & 58.28 & 83.99 & 49.43 & 67.70 & 76.80 & 58.98 \\
Meta-Llama-3-8B & 62.62 & 60.24 & 82.23 & 42.93 & 66.70 & 78.45 & 45.19 \\
\textbf{Bielik-4.5B-v3} & \textbf{61.02} & \textbf{51.19} & \textbf{73.01} & \textbf{45.63} & \textbf{61.32} & \textbf{71.35} & \textbf{63.61} \\
Mistral-7B-v0.1 & 60.97 & 59.98 & 83.31 & 42.15 & 64.16 & 78.37 & 37.83 \\
Mistral-7B-v0.2 & 60.37 & 60.84 & 83.08 & 41.76 & 63.62 & 78.22 & 34.72 \\
\textbf{Bielik-1.5B-v3} & \textbf{53.64} & \textbf{46.93} & \textbf{64.28} & \textbf{42.47} & \textbf{55.13} & \textbf{63.38} & \textbf{49.66} \\
Bielik-7B-v0.1 & 49.98 & 45.22 & 67.92 & 47.16 & 43.20 & 66.85 & 29.49 \\
\bottomrule
\end{tabular}
\caption{Open LLM Leaderboard results for base models}
\label{tab:open-llm-base}
\end{table*}

\begin{table*}[t]
\centering
\small
\begin{tabular}{lccccccc}
\toprule
\textbf{Model} & \textbf{AVG} & \textbf{arc\_challenge} & \textbf{hellaswag} & \textbf{truthfulqa\_mc2} & \textbf{mmlu} & \textbf{winogrande} & \textbf{gsm8k} \\
\midrule
SOLAR-10.7B-Instruct-v1.0 & 74.20 & 71.08 & 88.16 & 71.43 & 66.21 & 83.58 & 64.75 \\
Phi-3-medium-4k-instruct & 73.45 & 67.32 & 85.76 & 57.71 & 77.83 & 72.69 & 79.38 \\
Bielik-11B-v2.2-Instruct & 69.86 & 59.90 & 80.16 & 58.34 & 64.34 & 75.30 & 81.12 \\
Bielik-11B-v2.3-Instruct & 69.82 & 59.30 & 80.11 & 57.42 & 64.57 & 76.24 & 81.27 \\
Bielik-11B-v2.1-Instruct & 69.82 & 59.56 & 80.20 & 59.35 & 64.18 & 75.06 & 80.59 \\
openchat-3.5-0106-gemma & 69.42 & 64.68 & 81.08 & 54.93 & 64.69 & 78.30 & 72.86 \\
Bielik-11B-v2.0-Instruct & 68.04 & 58.62 & 78.65 & 54.65 & 63.71 & 76.32 & 76.27 \\
Meta-Llama-3-8B-Instruct & 66.87 & 60.75 & 78.55 & 51.65 & 67.07 & 74.51 & 68.69 \\
Mistral-7B-Instruct-v0.2 & 65.71 & 63.14 & 84.88 & 68.26 & 60.78 & 77.19 & 40.03 \\
\textbf{Bielik-4.5B-v3-Instruct} & \textbf{64.89} & \textbf{56.06} & \textbf{73.90} & \textbf{50.79} & \textbf{63.66} & \textbf{71.19} & \textbf{73.69} \\
gemma-7b & 64.29 & 61.09 & 82.47 & 44.91 & 66.03 & 78.45 & 52.77 \\
Qwen1.5-32B-Chat & 62.95 & 66.04 & 85.49 & 66.95 & 74.99 & 77.19 & 7.05 \\
Qwen1.5-14B-Chat & 62.27 & 58.70 & 82.27 & 60.36 & 68.57 & 73.09 & 30.63 \\
\textbf{Bielik-1.5B-v3-Instruct} & \textbf{56.64} & \textbf{48.38} & \textbf{65.03} & \textbf{42.47} & \textbf{54.59} & \textbf{65.35} & \textbf{62.85} \\
Qwen1.5-7B-Chat & 55.15 & 55.89 & 78.56 & 53.54 & 61.65 & 67.72 & 13.57 \\
Mistral-7B-Instruct-v0.1 & 54.96 & 54.52 & 75.63 & 56.28 & 55.38 & 73.72 & 14.25 \\
Bielik-7B-Instruct-v0.1 & 51.26 & 47.53 & 68.91 & 46.18 & 49.47 & 65.51 & 29.95 \\
\bottomrule
\end{tabular}
\caption{Open LLM Leaderboard results for selected instruction-tuned models}
\label{tab:open-llm-instruct}
\end{table*}

\subsection{MixEval}

MixEval \citep{ni2024mixeval} is an English-language benchmark grounded in verified data, created to assess Large Language Models (LLMs) both efficiently and reliably. Its main characteristics include:

\begin{enumerate}
    \item Built from a collection of pre-existing benchmark datasets

    \item Demonstrates strong alignment with Chatbot Arena rankings, showing a 0.96 correlation

    \item Executes locally with minimal overhead, requiring just 6\% of the time and cost of MMLU
\end{enumerate}
This benchmark offers a dependable and fast approach for evaluating LLMs, making it a practical choice for continuous performance tracking and model comparison. Results are presented in Table~\ref{tab:mixeval}.

\begin{table*}[t]
\centering
\small
\begin{tabular}{lcc}
\toprule
\textbf{Model} & \textbf{MixEval-Hard} & \textbf{MixEval} \\
\midrule
Qwen1.5-72B-Chat & 48.3 & 84.1 \\
LLaMA-3-8B-Instruct & 45.6 & 75.0 \\
Bielik-11B-v2.1-Instruct & 45.0 & 74.6 \\
Qwen1.5-32B-Chat & 43.3 & 81.0 \\
Bielik-11B-v2.3-Instruct & 43.2 & 73.0 \\
Bielik-11B-v2.0-Instruct & 40.2 & 72.1 \\
Bielik-11B-v2.2-Instruct & 39.7 & 72.4 \\
Mistral-7B-Instruct-v0.2 & 36.2 & 70.0 \\
\textbf{Bielik-4.5B-v3-Instruct} & \textbf{29.6} & \textbf{55.3} \\
\textbf{Bielik-1.5B-v3-Instruct} & \textbf{24.8} & \textbf{46.6} \\
\bottomrule
\end{tabular}
\caption{MixEval benchmark results comparing Bielik v3 models against other instruction-tuned models}
\label{tab:mixeval}
\end{table*}

\subsection{Berkeley Function-Calling Leaderboard}

The Berkeley Function-Calling Leaderboard (BFCL) \cite{berkeley-function-calling-leaderboard} is designed to measure the proficiency of language models in accurately invoking functions (tools) using realistic input data. This evaluation is critical for determining how effectively models can interact with APIs and external systems—an essential skill for deploying LLMs in areas such as software engineering, data processing, and automated workflows.

The benchmark employs Abstract Syntax Tree (AST) metrics to gauge function call correctness across a range of test types:
\begin{itemize}
    \item \textbf{Expert Curated (Non-live) dataset:} A collection of hand-crafted, static test cases developed by domain experts to assess function calling in controlled environments
    \item \textbf{User Contributed (Live) dataset:} Real-time, user-submitted examples that reflect function calling in authentic, dynamic situations
    \item \textbf{Multi-turn interactions:} Evaluates the model's capability to preserve and utilize conversational context over multiple exchanges
    \item \textbf{Relevance detection:} Determines whether the model appropriately triggers a function in cases where at least one relevant function should be used. Multiple valid function calls may exist; correctness of arguments is not strictly enforced—only that a relevant function is invoked
    \item \textbf{Irrelevance detection:} Tests the model's ability to refrain from invoking any functions when none are applicable. Models should either justify why no function is suitable or respond without initiating a function call
\end{itemize}

\begin{table*}[t]
\centering
\small
\begin{tabular}{lccccccccc}
\toprule
\textbf{Model} & \textbf{Non-Live} & \textbf{Non-Live} & \textbf{Non-Live} & \textbf{Non-Live} & \textbf{Live} & \textbf{Live} & \textbf{Live} & \textbf{Live Parallel} \\
 & \textbf{Python} & \textbf{Multiple} & \textbf{Parallel} & \textbf{Parallel} & \textbf{Simple} & \textbf{Multiple} & \textbf{Parallel} & \textbf{Multiple} \\
 & \textbf{Simple AST} & \textbf{AST} & \textbf{AST} & \textbf{Multiple AST} & \textbf{AST} & \textbf{AST} & \textbf{AST} & \textbf{AST} \\
\midrule
Open-Mistral-Nemo-2407 (Prompt) & 92.00\% & 93.50\% & 89.50\% & 84.50\% & 77.91\% & 74.45\% & 87.50\% & 66.67\% \\
Gemma-3-12b-it (Prompt) & 94.00\% & 95.00\% & 90.00\% & 73.00\% & 84.88\% & 70.85\% & 87.50\% & 62.50\% \\
Open-Mistral-Nemo-2407 (FC) & 91.25\% & 93.50\% & 85.50\% & 85.00\% & 77.13\% & 69.61\% & 75.00\% & 70.83\% \\
Bielik-11B-v2.5-Instruct (FC) & 95.00\% & 97.50\% & 87.50\% & 87.00\% & 77.13\% & 77.21\% & 43.75\% & 66.67\% \\ 
\textbf{Bielik-4.5B-v3.0-Instruct (FC)} & \textbf{94.00\%} & \textbf{92.50\%} & \textbf{82.00\%} & \textbf{86.00\%} & \textbf{70.16\%} & \textbf{68.66\%} & \textbf{50.00\%} & \textbf{54.17\%} \\ 
Qwen2.5-3B-Instruct (Prompt) & 91.50\% & 90.50\% & 79.50\% & 79.00\% & 69.77\% & 66.48\% & 56.25\% & 62.50\% \\
Qwen2.5-3B-Instruct (FC) & 96.00\% & 92.00\% & 73.50\% & 76.50\% & 74.03\% & 72.08\% & 62.50\% & 45.83\% \\
Qwen2.5-1.5B-Instruct (FC) & 92.25\% & 87.00\% & 81.50\% & 75.50\% & 74.03\% & 66.10\% & 50.00\% & 45.83\% \\
Qwen2.5-1.5B-Instruct (Prompt) & 89.00\% & 86.00\% & 70.00\% & 66.50\% & 70.54\% & 59.26\% & 56.25\% & 41.67\% \\
\textbf{Bielik-1.5B-v3.0-Instruct (FC)} &  \textbf{77.00\%} & \textbf{85.00\%} & \textbf{69.50\%} & \textbf{63.00\%} & \textbf{61.63\%} & \textbf{58.69\%} & \textbf{50.00\%} & \textbf{41.67\%} \\ 
Bielik-11B-v2.3-Instruct (Prompt) & 87.50\% & 93.50\% & 47.00\% & 50.00\% & 72.87\% & 69.71\% & 43.75\% & 54.17\% \\
\bottomrule
\end{tabular}
\caption{Comprehensive summary of model performance on the Berkeley Function-Calling Leaderboard subtasks. Bielik models demonstrate strong results across a variety of subtasks, excelling especially in Non-Live Python Simple AST and Non-Live Multiple AST categories, while also maintaining consistent outcomes in Live Simple and Multiple AST tasks.}
\label{tab:bfcl-subtasks}
\end{table*}

\section{Limitations and Biases}

Bielik v3 series of models can produce factually incorrect output, and should not be relied on to produce factually accurate data. Our models were trained on various public datasets. While great efforts have been taken to clear the training data, it is possible that this model can generate lewd, false, biased or otherwise offensive outputs.

\section{Conclusion}

In this technical report, we introduce the Bielik v3 series of generative text models for Polish language processing, including 1.5B and 4.5B parameter variants. These models represent a substantial advancement in Polish language AI, offering remarkable parameter efficiency while maintaining strong performance across diverse linguistic tasks, as demonstrated in our comprehensive evaluations in Tables~\ref{tab:open-pl-llm-base}, \ref{tab:open-pl-llm-instruct}, \ref{tab:pl-eq-bench}, \ref{tab:cptub}, and \ref{tab:medical-leaderboard}.

Key contributions of our work include:

\begin{enumerate}
    \item \textbf{Innovative Architectural Decisions}: Building upon the Qwen2.5 architecture, we implemented depth up-scaling and replaced the tokenizer with our custom APT4 tokenizer optimized for Polish, resulting in more efficient token usage.

    \item \textbf{Data Quality Focus}: We developed sophisticated quality classification systems with 95\% accuracy to ensure our training corpus consisted of high-quality Polish texts balanced across 120 thematic categories.

    \item \textbf{Training Methodology Innovations}: Our techniques include Adaptive Learning Rate, which significantly improved model performance, particularly for Polish-specific linguistic patterns.

    \item \textbf{Impressive Performance Efficiency}: The 4.5B parameter model achieves results competitive with models 2-3$\times$ larger across multiple benchmarks, while the 1.5B model delivers strong performance despite its extremely compact profile. This efficiency makes Bielik v3 particularly valuable for deployment in resource-constrained environments while maintaining high-quality Polish language capabilities.

    \item \textbf{Benchmark Excellence}: On the Open PL LLM Leaderboard, CPTUB, Polish Medical Benchmark, and EQ-Bench, Bielik v3 models consistently outperform many larger models, demonstrating exceptional efficiency.
\end{enumerate}

These models provide a powerful foundation for Polish language applications across various domains, from general conversational AI to specialized fields such as medicine and law. By prioritizing parameter efficiency without sacrificing quality, Bielik v3 enables broader deployment on resource-constrained systems while advancing the state of Polish language AI.

Future work will focus on further enhancing capabilities for complex reasoning, exploring additional efficiency improvements, and expanding domain-specific knowledge. We believe the Bielik v3 models establish a new benchmark for efficient, high-quality language models for less-resourced languages.

\section*{Acknowledgements}

We gratefully acknowledge Polish high-performance computing infrastructure PLGrid (HPC Center: ACK Cyfronet AGH) for providing computer facilities and support within computational grant no. PLG/2024/017214 and PLG/2025/018338.

The model could not have been created without the commitment and work of the entire SpeakLeash team, whose contribution is invaluable. Thanks to the hard work of many individuals, it was possible to gather a large amount of content in Polish and establish collaboration between the open-science SpeakLeash project and the HPC center: ACK Cyfronet AGH. Individuals who contributed to the creation of the model through their commitment to the open-science SpeakLeash project: Sebastian Kondracki, Marek Magryś, Szymon Mazurek, Mieszko Cholewa, Igor Ciuciura, Szymon Baczyński, Jacek Chwiła, Dominika Basaj, Kuba Sołtys, Karol Jezierski, Anna Przybył, Agnieszka Ratajska, Witold Wydmański, Izabela Babis, Nina Babis, and many other wonderful researchers and enthusiasts of the AI world.

\bibliographystyle{unsrtnat}
\bibliography{references} 

\end{document}